\documentclass[lettersize,journal]{IEEEtran}
\usepackage{amsmath,amsfonts}
\usepackage{algorithmic}
\usepackage{array}
\usepackage[caption=false,font=footnotesize,labelfont=rm,textfont=rm]{subfig}
\usepackage{textcomp}
\usepackage{stfloats}
\usepackage{url}
\usepackage{verbatim}
\usepackage{graphicx}

\usepackage[table,xcdraw]{xcolor}
\definecolor{GREEN}{RGB}{175, 211, 153}
\definecolor{RED}{RGB}{255, 190, 190}
\definecolor{BLUE}{RGB}{28, 152, 214}
\newcommand{\tabincell}[2]{\begin{tabular}{@{}#1@{}}#2\end{tabular}}
\usepackage{multirow}
\usepackage[ruled]{algorithm2e}
\usepackage{setspace}
\usepackage{makecell}  
\usepackage{amsmath}
\usepackage{amssymb}  
\usepackage{subfig}
\usepackage{float}
\usepackage{booktabs}

\usepackage{balance}
\usepackage{hyperref}

\begin{document}
\title{Self-similarity Prior Distillation for Unsupervised Remote Physiological Measurement}
\author{Xinyu Zhang, Weiyu Sun, Hao Lu, Ying Chen, Yun Ge, Xiaolin Huang, Jie Yuan, Yingcong Chen

\thanks{Xinyu Zhang, Weiyu Sun, Ying Chen, Yun Ge, Xiaolin Huang, and Jie Yuan are with the School of Electronic Science and Engineering, Nanjing University, Nanjing 210023, China. E-mail: \{xinyuzhang, weiyusun\}@smail.nju.edu.cn, \{yingchen, geyun, xlhuang, yuanjie\}@nju.edu.cn.}
\thanks{Hao Lu and Yingcong Chen are with the Artificial Intelligence Thrust, Hong Kong University of Science and Technology, Guangzhou 511453, China. E-mail: hlu585@connect.hkust-gz.edu.cn, yingcongchen@ust.hk.}
\thanks{The code is available at https://github.com/LinXi1C/SSPD}
}

\markboth{Journal of \LaTeX\ Class Files,~Vol.~18, No.~9, September~2020}%
{How to Use the IEEEtran \LaTeX \ Templates}

\maketitle

\begin{abstract}
Remote photoplethysmography (rPPG) is a non-invasive technique that aims to capture subtle variations in facial pixels caused by changes in blood volume resulting from cardiac activities. Most existing unsupervised methods for rPPG tasks focus on the contrastive learning between samples while neglecting the inherent self-similarity prior in physiological signals. In this paper, we propose a Self-Similarity Prior Distillation (SSPD) framework for unsupervised rPPG estimation, which capitalizes on the intrinsic temporal self-similarity of cardiac activities. 
Specifically, we first introduce a physical-prior embedded augmentation technique to mitigate the effect of various types of noise. Then, we tailor a self-similarity-aware network to disentangle more reliable self-similar physiological features. Finally, we develop a hierarchical self-distillation paradigm for self-similarity-aware learning and rPPG signal decoupling. Comprehensive experiments demonstrate that the unsupervised SSPD framework achieves comparable or even superior performance compared to the state-of-the-art supervised methods. Meanwhile, SSPD has the lowest inference time and computation cost among end-to-end models. 
\end{abstract}


\section{Introduction}
Cardiac activity is a fundamental physiological mechanism in the human body, and the assessment of physiological indicators, such as heart rate (HR), plays a critical role in maintaining physical and mental health. Traditionally, the acquisition of physiological measurements relies on contact medical instruments to obtain electrocardiogram (ECG) or photoplethysmography (PPG) signals, which may induce discomfort to users. Furthermore, in certain scenarios, such as driving \cite{drive2021, reddy2019autoencoding} or telemonitoring \cite{cheng2021}, contact devices are no longer convenient. The remote photoplethysmography (rPPG) method, which captures subtle facial pixel changes caused by blood volume variations, has emerged as a promising alternative due to its non-invasive and low-cost characteristics.

In recent years, rPPG methods \cite{Dual-GAN, PhysFormer, Contrast-Phys} have exhibited competitive performance for physiological measurement under the prevalence of deep learning. The primary challenges in rPPG tasks are related to the feebleness of the physiological signals and the strong noise in video recordings, including head movements \cite{Deepphys}, illumination changes \cite{RhythmNet}, and video artifacts \cite{rPPGnet}. To tackle these issues, researchers have elaborated various rPPG datasets \cite{VIPL, NIRP, HCI} and proposed effective supervised learning models \cite{CAN, PhysNet, SynRhythm} for rPPG estimation. The acquisition of high-quality labels requires high-precision contact instruments, such as electrodes for ECG \cite{HCI} and photoelectric sensors for BVP \cite{VIPL, NIRP, PURE, UBFC}. Furthermore, the signals must be recorded for a long period with contact, which can be a labor-intensive and time-consuming process \cite{fewshot2022}. Fortunately, unsupervised methods are a promising alternative as they can free the model from the dependence on supervision signals.

Unsupervised methods in the rPPG field can be categorized into two branches: signal-based and learning-based methods. Traditional signal-based methods \cite{ICA, CHROM, GREEN, POS} tend to design models manually according to some specific assumptions (e.g., skin reflection model \cite{POS}). These assumptions are mere simplifications of real scenarios, thus making signal-based methods vulnerable to various environment noise. On the other hand, most existing learning-based methods tend to use contrastive learning to extract physiological features, focusing on how to design positive and negative sample pairs,
which can be generated from the same clip with temporal disordering \cite{Self-rPPG} or frequency resampling \cite{RemotePPG, yue2022}, or from different clips with augmentation \cite{SLF-RPM, ViViT} or frequency resampling \cite{ContrastPhys2}. The contrastive loss can be performed on the latent representations followed by fine-tuning \cite{Self-rPPG, SLF-RPM, ViViT}, directly on the output rPPG \cite{yue2022}, or on the Power Spectrum Density (PSD) derived from the rPPG \cite{Contrast-Phys, RemotePPG, ContrastPhys2}.

However, these methods ignore the crucial self-similarity prior inherent in cardiac activities. Self-similarity indicates consistency in a dynamic system where a pattern or event appears similar to itself across different scales or intervals.
In remote physiological measurement, self-similarity can be observed from recurring actions in every cardiac cycle, such as ventricular contraction and relaxation \cite{rouast2018remote}, 
resulting in a periodic heart rate. This periodicity further leads to self-similar patterns appearing at different time intervals in both facial video and rPPG signal. 
Compared to the instance-level prior used in contrastive learning \cite{Contrast-Phys, RemotePPG, SLF-RPM}, self-similarity has a finer granularity.
Specifically, contrastive learning emphasizes the distance metric between positive and negative sample pairs, while self-similarity prior distillation focuses on effectively capturing the physiological features within individual samples.

In this paper, we propose a self-similarity prior distillation framework for unsupervised remote physiological measurement, which capitalizes on the intrinsic self-similarity of cardiac activities. First, we develop a physical-prior embedded augmentation technique, consisting of Local-Global Augmentation (LGA) and Masked Difference Modeling (MDM) against noise from both spatial and temporal aspects. Next, we tailor a self-similarity-aware network that comprises a backbone, a predictor module, and a Separable Self-Similarity Model ($S^3M$). The predictor module is responsible for estimating the rPPG signal, while the $S^3M$ generates a temporal similarity pyramid consisting of self-similarity maps at multiple time scales, facilitating self-similarity-aware learning. Furthermore, $S^3M$ serves as a training auxiliary to boost the performance without incurring extra computational expenses during inference. Finally, we design a hierarchical self-distillation paradigm to help the network disentangle self-similar physiological patterns from facial videos. 
The proposed hierarchical self-distillation involves Temporal Similarity Pyramid Distillation (TSPD) and RPPG Prediction Distillation (RPD) for self-similarity-aware learning and rPPG signal decoupling, respectively. 

In summary, the contributions of this work are as follows:

(1) We propose a Self-Similarity Prior Distillation (SSPD) framework for unsupervised remote physiological measurement, which exploits the inherent self-similarity of cardiac activities via a hierarchical self-distillation strategy.

(2) We develop two physical-prior embedded augmentation strategies to mitigate the effect of various types of noise from both spatial and temporal aspects. 

(3) We devise a Separable Self-Similarity Model ($S^3M$) to enable self-similarity-aware learning, forcing the network to learn multi-scale and long-distance physiological features better, without any extra computing expenses in the inference stage.

(4) We conduct extensive evaluations on four commonly used open-access benchmarks (PURE \cite{PURE}, UBFC-rPPG \cite{UBFC}, VIPL-HR \cite{VIPL}, and MR-NIRP \cite{NIRP}). The substantial experimental results demonstrate that SSPD outperforms the state-of-the-art unsupervised methods significantly.

\section{Related Works}
\subsection{Self-supervised Learning}
Unsupervised/self-supervised learning aims to let model learn a good representation from input data without human-annotated labels, guided by solving the pretext task. In recent years, self-supervised learning has achieved great success in natural
language processing \cite{BERT, GPT} and computer vision tasks \cite{MoCo, MAE, DINO}. Self-supervised methods presented to date can be broadly categorized into three paradigms: contrastive learning \cite{MoCo, Siamese, SimCLR, SimCSE}, masked modeling \cite{BERT, MAE, Beit, MSN} and self-distillation \cite{DINO, iBOT}. The fundamental principle of contrastive learning is to attract positive sample pairs while repulsing negative pairs \cite{contrast_learning}. However, BYOL \cite{BYOL} demonstrates that contrastive learning can still be effective without the use of negative samples. Masked modeling performs a pretext task of predicting the masked content based on the remaining uncorrupted input. In other words, the supervision signal for masked modeling is derived solely from the input samples themselves. Self-distillation is a type of knowledge distillation \cite{knowledge_Distillation, CDFKD-MFS} in which the model learns from itself. Specifically, the teacher and student models share an identical structure, and the output of different layers can serve as supervision signals, which are also known as distilled knowledge. To obtain better knowledge, different instance discrimination \cite{Instance_discrimination} strategies have been proposed to generate diverse input views. Caron et al. \cite{DINO} presented a self-distillation model DINO, leveraging the multi-crop \cite{SWaV} strategy to promote local-to-global correspondences. Furthermore, Zhou et al. \cite{iBOT} proposed the iBOT model, which utilizes masked image modeling inspired by BERT \cite{BERT}.
Unlike the conventional self-supervised learning paradigm of unsupervised pre-training followed by supervised fine-tuning, 
we can unsupervisedly decouple the periodic physiological signals directly without fine-tuning. Therefore, the whole process is label-free.

\subsection{Remote Physiological Measurement}
Traditional methods take the approach of decomposing the temporal sequence signal to expose the rPPG signal of genuine interest. Therefore, signal disentanglement methods (e.g., ICA \cite{ICA}, PCA \cite{PCA, yang2017PCA}) and skin reflection modeling methods (e.g., CHROM \cite{CHROM}, POS \cite{POS}) are proposed. However, deep learning methods take the path of estimating the rPPG and noise distribution from the input video and thus have better robustness. In supervised rPPG estimation, Chen et al. \cite{Deepphys} proposed the first end-to-end model, namely DeepPhys. This model takes the original frame and frame difference as inputs to the appearance and motion models, respectively, fusing them through an attention mechanism. Building upon this, the two-stream models \cite{CAN, Two-stream, IAN} have been demonstrated to be effective for rPPG estimation. Meanwhile, spatial-temporal representation learning \cite{rPPGnet, PhysNet, TransPPG, pain2021} has emerged as a promising alternative, which typically employs a 3D convolutional neural network (CNN) or vision transformer \cite{ViT} as the backbone. Additionally, an effective preprocessing paradigm called STMap \cite{Dual-GAN, RhythmNet, CVD} has been designed. STMap involves the selection of multiple regions of interest (ROIs) from facial video, which allows the network to focus more on rPPG features and reduce the impact of noise. 

However, annotating large-scale datasets in deep learning \cite{Imagenet, K700}, particularly in physiological measurement, has always been a labor-intensive and time-consuming process. The acquisition of long-period physiological signals as ground truth is expensive and challenging, which has led to the emergence of self-supervised learning as a promising alternative for rPPG estimation \cite{Contrast-Phys, RemotePPG, SLF-RPM}. Gideon and Stent \cite{RemotePPG} leveraged the smooth heart rate prior that HR ranges between 40 to 250 bpm and HR is stable over short time intervals. They adopted spectrum perturbation to generate positive-anchor-negative triplet samples for contrastive learning. Furthermore, Sun and Li \cite{Contrast-Phys} proposed the Contrast-Phys method, which utilized spatial and temporal similarity to generate positive sample pairs. Additionally, Park et al. \cite{ViViT} presented a contrastive vision transformer framework for the joint modeling of RGB and near-infrared (NIR) videos.
Most existing self-supervised methods focus on constructing positive and negative samples for contrastive learning. 
Unlike extracting instance-level features, our SSPD framework identifies the more fine-grained self-similar patterns to capture temporal rhythms synchronized with the periodic heart rate within individual samples.

\section{Method}
\subsection{Overview}
In this section, we initially provide a detailed explanation of the self-similarity prior in rPPG. We introduce Self-Similarity-Map (SSM) to assist the model in focusing on temporal self-similarity and propose Self-Similarity-Wave (SSW) to enhance the learning of the inherent periodicity in SSM. Subsequently, we present the SSPD framework to achieve self-similarity-aware learning and rPPG signal decoupling through hierarchical self-distillation. Additionally, we integrate two periodicity regularizations derived from the self-similarity prior to help the model better perceive periodicity, facilitating the construction of more reliable SSM and SSW.

\subsection{Self-similarity Prior in rPPG}\label{Self-similarity Prior in rPPG}
Self-similarity is a crucial physiological phenomenon that describes the recurring actions in every cardiac cycle associated with heartbeat generation \cite{rouast2018remote}.
This cyclic behavior results in the recurrence of distinct patterns within the rPPG signal over time, 
such as falling and rising edges. To facilitate learning self-similar features in facial videos that synchronize with cardiac rhythms, 
we propose Self-Similarity Map (SSM) and Self-Similarity Wave (SSW), modeling the intrinsic self-similarity prior for unsupervised physiological measurement.

\begin{figure}[H]
    \vspace{-0.5cm}
    \centering
    \subfloat[rPPG signal]{\includegraphics[height=2.8cm]{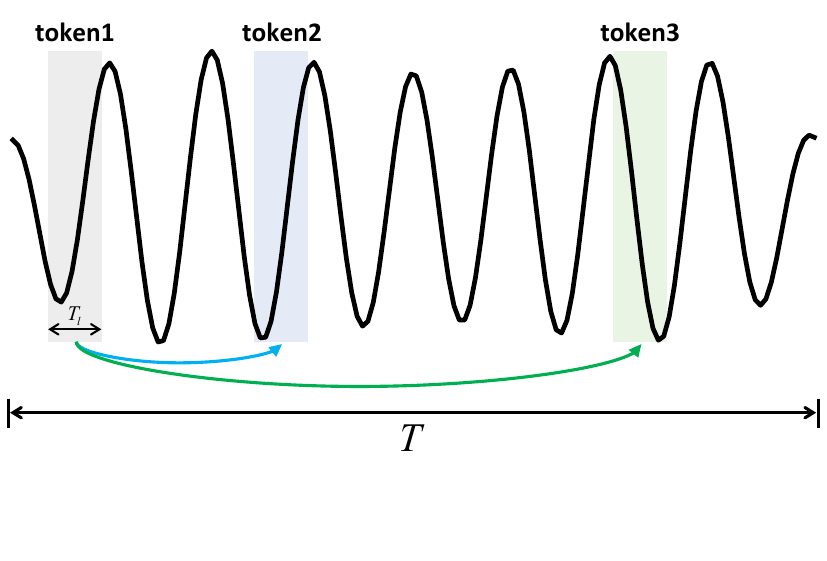}}\hspace{0.3cm}
    \subfloat[SSM: Self-Similarity Map]{\includegraphics[height=3.5cm]{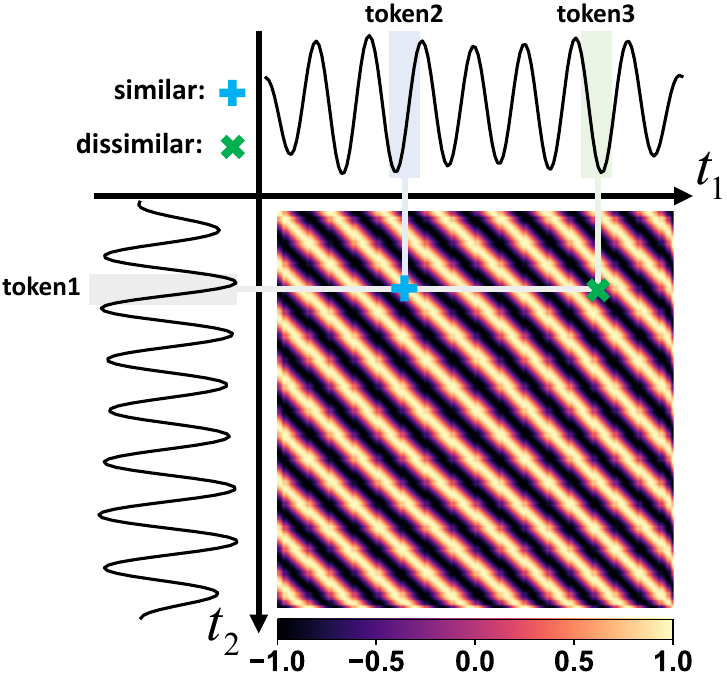}}
    \caption{Self-similarity map characterizes the self-similarity of each token pair in rPPG signals.}
    \label{rPPG_SSM}
\end{figure}

\subsubsection{Self-Similarity Map (SSM)}
Self-similarity originates from the periodic heart rate, resulting in self-similar patterns at different time intervals in the periodic rPPG signal.
To direct the model's focus on self-similarity over other irrelevant features, 
we convert the output feature map into a Self-Similarity Map (SSM), representing the similarity between patterns at each time interval.
As depicted in Fig. \ref{rPPG_SSM}(a), we define each time window of length $T_l$ and stride 1 within the rPPG signal, 
which has a length of $T$, as a ``token'' denoted as $u_i$. Each token symbolizes a specific action within cardiac cycles. 
We iterate over all tokens $\{u_1, u_2, .., u_{T-T_l+1}\}$, $u_i \in \mathbb{R}^{T_l}$, and compute the cosine similarity between each pair of tokens ($u_i$, $u_j$) to construct the self-similarity map $\mathcal{M} \in \mathbb{R}^{(T-T_l+1) \times (T-T_l+1)}$ as follows:

\begin{equation}
    \begin{aligned}
        \mathcal{M}_{ij} &= Sim(u_i, u_j) = \frac{{u_i} \cdot u_j}{\Vert u_i \Vert_2 \cdot \Vert u_j \Vert_2}
    \end{aligned}
    \label{SSM}
\end{equation}

In Fig. \ref{rPPG_SSM}(b), token1 is similar to token2 and dissimilar to token3, 
corresponding to SSM's high and low values. In contrast to the rPPG signal, 
SSM concentrates explicitly on the temporal similarity of each token pair. 
This focus allows SSM to filter out attributes less crucial to heartbeat rhythms, 
such as amplitude magnitude and envelope shape. 
Also, SSM provides a fine-grained contextual relationship. 
By forming SSM from the similarity of tokens at different time intervals, 
the network can learn long-distance physiological features better, 
serving as the ``self-similarity-based attention'' mechanism \cite{Transformer}.

Noteworthy is that SSM has two inherent properties. Firstly, SSM is a symmetric matrix, i.e., $\mathcal{M}_{ij} = \mathcal{M}_{ji}$. Thus, we typically analyze the upper triangular matrix to avoid redundancy unless otherwise indicated. 
Secondly, SSM has a main diagonal of 1 and $\mathcal{M}_{ij} \in [-1, 1]$.

\begin{figure}[H]
    \centering
    \includegraphics[height=9cm]{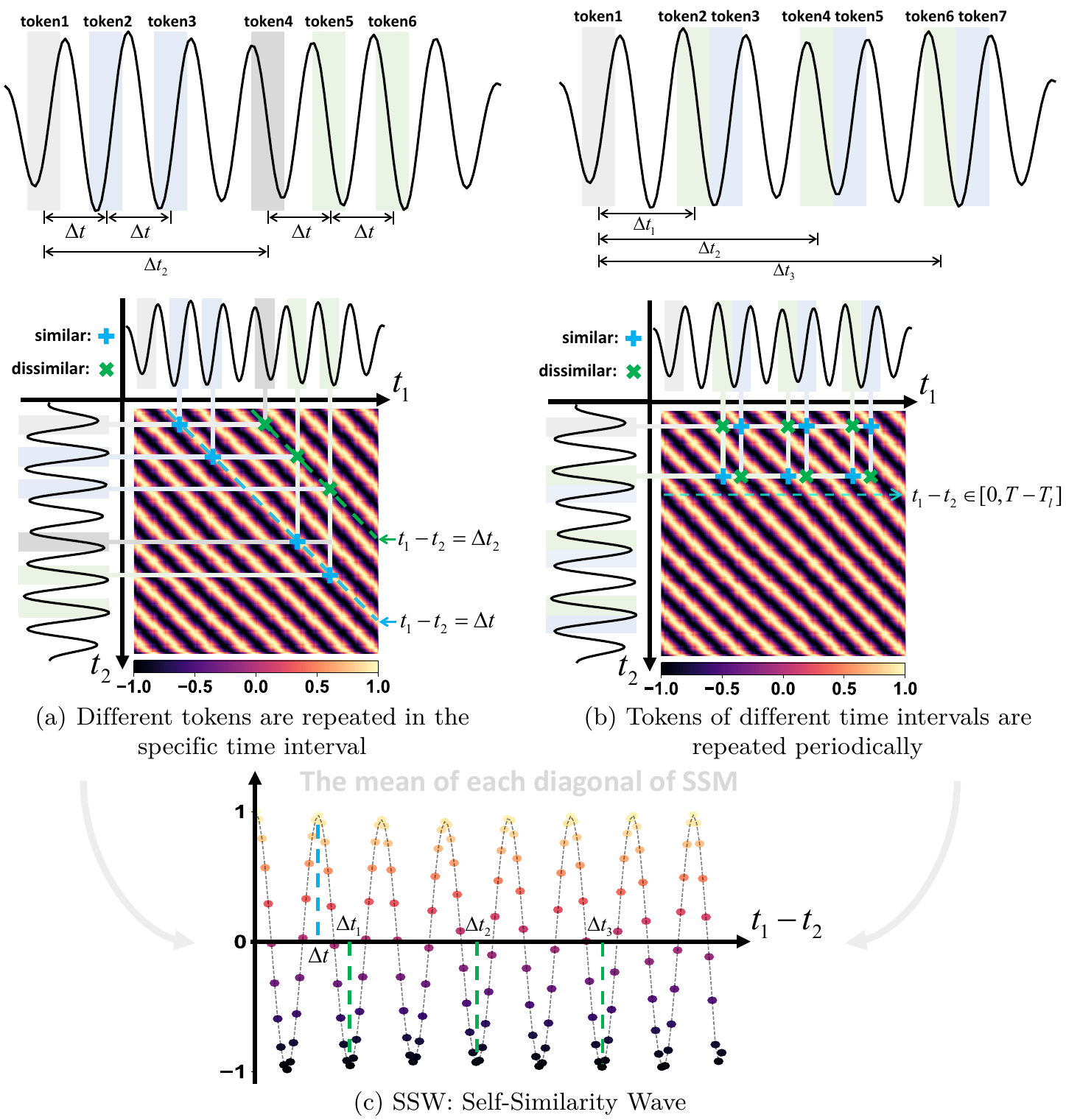}
    \caption{Interrelationships between rPPG signal, self-similarity map, and self-similarity wave. Based on prior knowledge, the self-similarity wave extracts the strong periodic component from the self-similarity map.}
    \label{SSM_SSW}
\end{figure}

\subsubsection{Self-Similarity Wave (SSW)}
Although we have constructed SSM to characterize self-similarity, 
its fine-grained and long-distance nature also makes it challenging for the network to learn reliable SSM without supervision directly.
To facilitate SSM learning, we refine the Self-Similarity Wave (SSW) from SSM to exploit periodicity, 
leveraging insights from the smooth heart rate prior introduced in previous works \cite{Contrast-Phys, RemotePPG}. 
The smooth heart rate prior implies that the heart rate remains relatively stable within a certain time window, 
typically lasting less than 10 seconds. Under this assumption, we can exploit periodicity from SSM to better capture self-similarity.

First, with a stable heat rate, \emph{different tokens are repeated in a specific time interval}.
As shown in Fig. \ref{SSM_SSW}(a), token1 exhibits the highest similarity with token2, 
and token2 shares a similar pattern with token3 within the same interval $\Delta t$. 
This phenomenon is also evidenced between token4, token5, and token6. 
Corresponding to the SSM, we observe that the similarities of these token pairs are located on the same diagonal of SSM, 
denoted as $d_{\Delta t}=\{Sim(u_i, u_j) | i-j = \Delta t \}$. Conversely, token1 is dissimilar to token4, 
and the same dissimilarity is observed between token2 and token5, 
as well as token3 and token6, all sharing the same time interval $\Delta t_2$. 
Also, these dissimilarities are located on the same diagonal $d_{\Delta t_2}=\{Sim(u_i, u_j) | i-j = \Delta t_2 \}$.
Overall, different diagonals $\{d_1, d_2, .., d_{T-T_l+1}\}$, $d_i \in \mathbb{R}^{T-T_l+2-i}$, in SSM signify the similarities of token pairs with specific time intervals, 
with each diagonal comprising nearly identical elements attributed to heart rate periodicity.

\begin{figure*}[t]
    \centering
    \includegraphics[height=8.5cm]{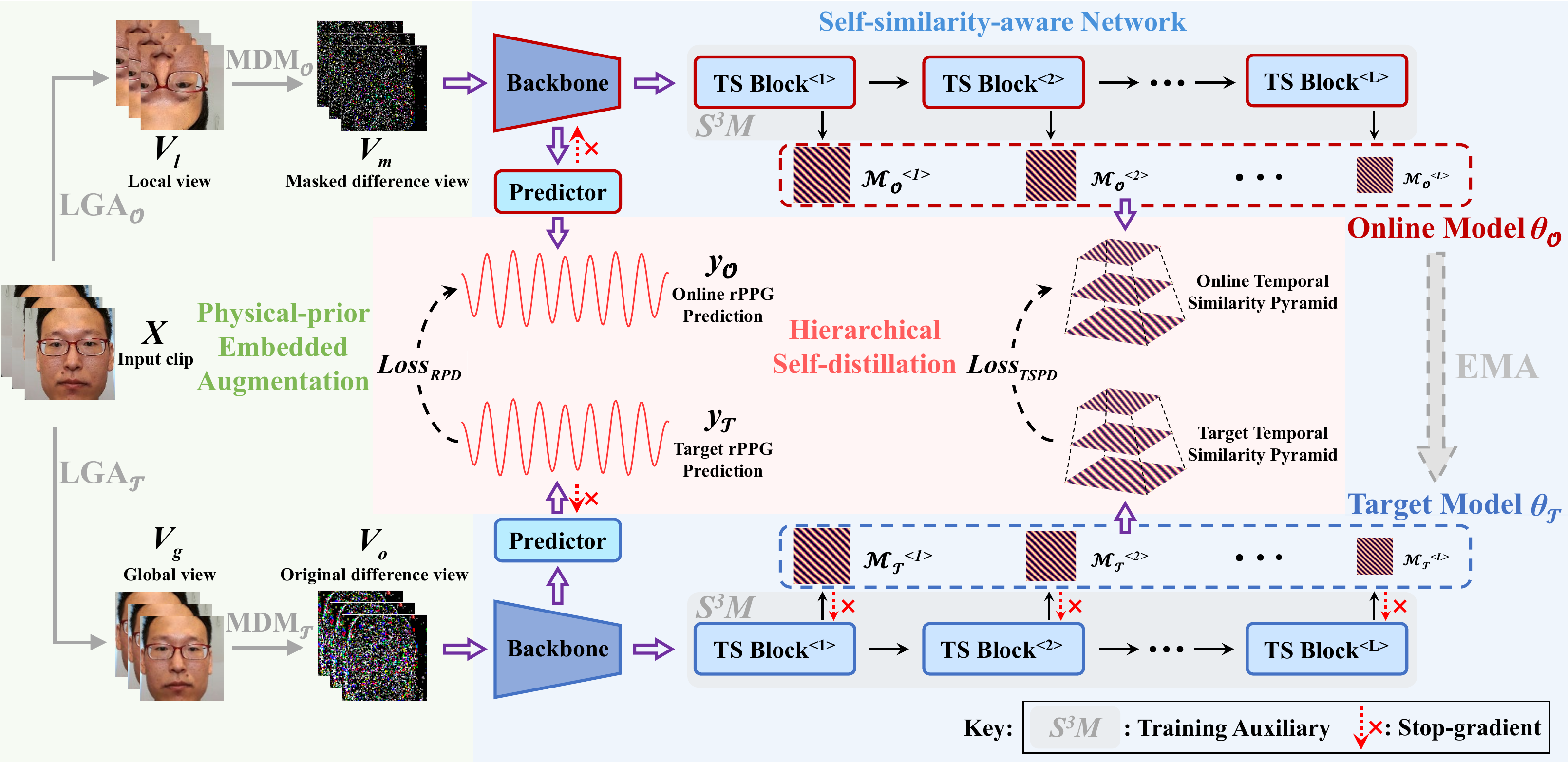}
    \caption{The architecture of our SSPD framework for unsupervised remote physiological measurement. This framework first incorporates Local-Global Augmentation (LGA) and Masked Difference Modeling (MDM) to generate two distorted views. Then, the tailored self-similarity-aware network consists of a backbone, a predictor module, and a Separable Self-Similarity Model ($S^3M$). Each Temporal Similarity (TS) block in $S^3M$ exploits self-similar representations at a specific time scale, forming the temporal similarity pyramid $\{\mathcal{M}^{<1>},\mathcal{M}^{<2>},...,\mathcal{M}^{<L>}\}$, 
    and the $S^3M$ is used exclusively for training.
    Finally, the proposed hierarchical self-distillation paradigm comprises Temporal Similarity Pyramid Distillation (TSPD) and RPPG Prediction Distillation (RPD), 
    enabling self-similarity-aware learning and rPPG signal decoupling.}
    \label{SSPD}
\end{figure*}

Besides, \emph{tokens of different time intervals are repeated periodically}, as illustrated in Fig. \ref{SSM_SSW}(b). 
Taking token1 as an example, token2, token3, and token4 with the same similarity are repeated periodically. 
Similarly, token3, token5, and token6 also exhibit periodic repetition.
This phenomenon is evident in every row of the SSM, where the similarity across different time intervals shows a periodic change. 
If we idealize rPPG as a periodic signal, then the period of each row in SSM is the heart rate.
Building upon the prior knowledge in Fig. \ref{SSM_SSW}, 
we formulate the Self-Similarity Wave (SSW) $\mathcal{W} \in \mathbb{R}^{T-T_l+1}$ by extracting the mean of each diagonal of the SSM as follows:

\begin{equation}
    \begin{aligned}
        \mathcal{W}_i = \frac{1}{T-T_l+2-i}\sum_{n=1}^{T-T_l+2-i} d_{i,n} 
    \end{aligned}
    \label{SSW1}
\end{equation}

Each value in SSW represents the mean similarity of all token pairs with a specific time interval, with the highest similarity at $\Delta t$ and the lowest at $\Delta t_1$,  $\Delta t_2$, and $\Delta t_3$ in Fig. \ref{SSM_SSW}. 
In contrast to SSM, SSW introduces the smooth heart rate prior, extracting the strong periodic component from SSM. 
This periodicity helps the network perceive self-similarity better and extract more reliable physiological features than directly learning fine-grained SSM, especially in the early training stage.

\subsubsection{Self-similarity Prior with Unsupervised Learning}
Our motivation is to capture temporal self-similarity synchronized with the periodic heart rate among tokens (as illustrated in Fig. \ref{SSM}) derived from facial videos, without relying on ground truth. 
To enable unsupervised self-similarity-aware learning and rPPG signal decoupling in the SSPD framework, 
we draw inspiration from superior self-distillation frameworks \cite{DINO, iBOT} to distill shared self-similarity knowledge from past iterations of the model (i.e., target model) to the updated model (i.e., online model) between two distorted views. 
In this process, we design a learnable tokenizer to generate tokens from facial videos to formulate SSM and SSW for self-similarity-aware learning.
Moreover, since the input facial video shares the same self-similarity with the predicted rPPG, 
we can decouple the rPPG signal directly from the trained self-similar representations without the need to fine-tune it with the BVP label. 
In the subsequent sections, we will describe how the SSPD framework leverages the self-similarity prior to achieving unsupervised physiological measurements, as presented in Fig. \ref{SSPD}.

\subsection{Physical-prior Embedded Augmentation}\label{Physical-prior embedded augmentation}
In remote physiological measurement, noise sources can be broadly categorized into spatial and temporal noise, 
including head movements \cite{Deepphys}, illumination changes \cite{RhythmNet}, and video artifacts \cite{rPPGnet}. 
To capture shift-invariant features in the space domain, 
we introduce the Local-Global Augmentation (LGA) method to address spatial noise. 
Furthermore, we employ Masked Difference Modeling (MDM) to help the model focus more on motion-related information, 
mitigate the impact of temporal noise, and facilitate self-similarity-aware learning.

\subsubsection{Video Preprocessing}
Initially, the facial region is detected and cropped from the original video clip with a length of $T+1$ using the Dlib library \cite{Dlib}. The bounding box is only generated based on the first frame and maintained in the subsequent frames of the clip. Next, the bounding box is resized to 151$\times$151 for physical-prior embedded augmentation, denoted as 
$X\in\mathbb{R}^{(T+1)\times3\times151\times151}$.

\subsubsection{Local-global Augmentation}
We employ Local-Global Augmentation (LGA) for spatial augmentation. The LGA strategy aims to enable the model to capture shift-invariant features by enhancing local-global responses. This makes the model more robust to spatial variations in the facial area, such as head movements. 
Specifically, we generate the local view $V_l\in\mathbb{R}^{(T+1)\times3\times128\times128}$ and the global view 
$V_g\in\mathbb{R}^{(T+1)\times3\times128\times128}$ from the same input clip $X$.
The local view $V_l$ is produced by random resized cropping, random horizontal flipping, and random adding Gaussian noise with the shape of 128$\times$128, while the global view $V_g$ is obtained by resizing, randomly horizontal flipping with the same shape. Noteworthy, the augmentations utilized in LGA are solely applied to the spatial dimension, indicating that each frame in the video clip undergoes identical transformations. 


\subsubsection{Masked Difference Modeling}
Building upon the aforementioned LGA strategy, we elaborate the Masked Difference Modeling (MDM) for temporal augmentation, which leverages the self-similarity prior of physiological signals. The procedure of MDM can be described as 
follows:

\begin{equation}
    \begin{aligned}
        V_m &= m_p \odot (\Delta_t V_{l,t}) \\
        V_o &= \Delta_t V_{g,t}
    \end{aligned}
    \label{LGA and MDM}
\end{equation}

\noindent specifically, we calculate the first forward difference \cite{Deepphys} between two consecutive frames of $V_l$, where $\Delta_tV_{l,t}=V_{l,t+1}-V_{l,t}$. 
Then, we apply a binary mask $m_p\in\{0,1\}^{T\times3\times128\times128}$ to $\Delta_tV_{l,t}$, where each element of $m_p$ is independently sampled from the Bernoulli distribution, taking the value 0 with probability $p$.
The resulting masked difference view $V_m$ is the element-wise multiplication of the mask and the difference noted as $\odot$. 
Besides, the original difference view $V_o$ is solely generated through frame difference without masking. 

The reason why we apply a mask to the frame difference is that the mask itself can be viewed as a nontrivial pretext task \cite{MAE}. 
The presence of self-similar patterns in physiological signals enables the reconstruction of features affected by the mask from other frames at different timestamps, 
which promotes the model's ability to capture temporal self-similar representations.
Also, the mask ratio is always dynamic since the frame difference of static pixels in the video is 0, 
while a dynamic data augmentation technique can further improve the robustness of the model.

\begin{figure}
    \centering
    \includegraphics[height=8.0cm]{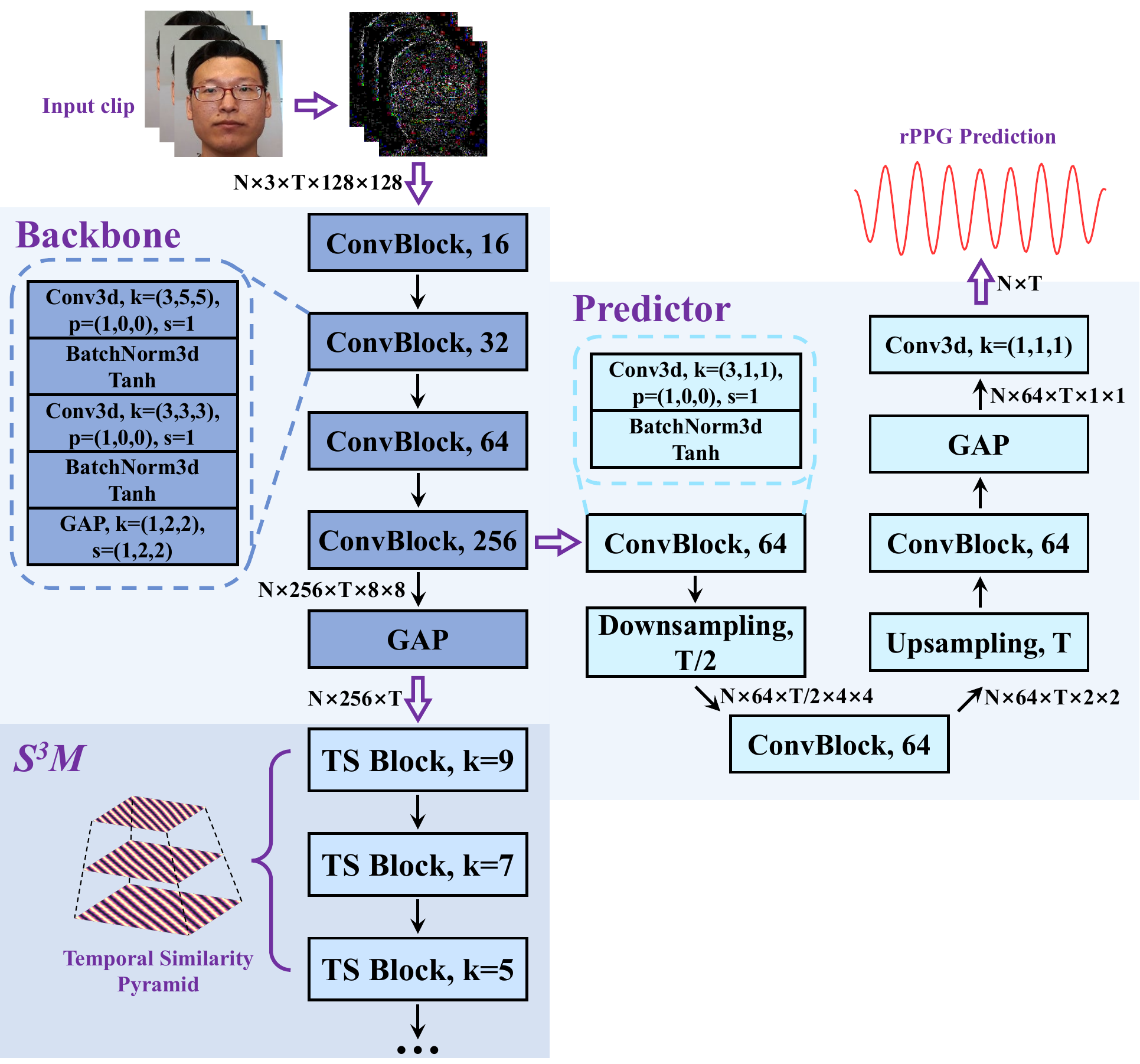}
    \caption{The architecture of the self-similarity-aware network. Each ``ConvBlock'' comprises two convolution layers, with ``GAP'' representing global average pooling. ``Downsampling'' refers to a reduction of half in the temporal domain. The dimensions are presented as N$\times$C$\times$T$\times$H$\times$W, where N and C indicate the number of samples and channels, respectively.}
    \label{backbone predictor module}
\end{figure}

\subsection{Self-similarity-aware Network}\label{Self-similarity-aware Network}
Self-similarity prior indicates recurring actions in every cardiac cycle, resulting in similar patterns at different intervals in both facial video and rPPG signal. 
To capture self-similar representations synchronized with heartbeat rhythms, 
we present a self-similarity-aware network consisting of a backbone for spatial-temporal feature extraction, 
a predictor module for rPPG signal decoupling, 
and a Separable Self-Similarity Model ($S^3M$) for self-similarity-aware learning.

\subsubsection{Backbone \& Predictor Module}
The backbone in the SSPD framework consists of four spatial-temporal convolutional blocks, which are designed to produce preliminary spatial-temporal feature maps. To decouple the rPPG signal, a predictor module is attached before the last Global Average Pooling (GAP) layer in the backbone, and the detailed structure of the backbone and predictor module are presented in Fig. \ref{backbone predictor module}. During the training and inference stages, 
the backbone of the online model $\theta_\mathcal{O}$ and target model $\theta_\mathcal{T}$ receive the masked difference view $V_m$ and the original difference view $V_o$ as input, respectively.
The outputs of the backbones are represented as $S_\mathcal{O}\in\mathbb{R}^{T\times C}$ and $S_\mathcal{T}\in\mathbb{R}^{T\times C}$ , and the predicted rPPG output from the predictor module can be expressed as $y_\mathcal{O}\in\mathbb{R}^T$ and $y_\mathcal{T}\in\mathbb{R}^T$, respectively. 

Unlike the two-stage strategy of unsupervised pre-training followed by supervised fine-tuning in conventional self-supervised frameworks \cite{SLF-RPM, ViViT, MoCo, MAE, DINO}, 
our SSPD exploits the property of shared self-similarity between the rPPG signal and the video signal to directly decouple the rPPG signal from self-similar representation. 
Specifically, the backbone and $S^3M$ are responsible for learning self-similarity from facial videos, while the predictor module constructs rPPG waveforms based on the trained self-similar representations. 
However, we cannot guarantee that the predictor can accurately decouple rPPG waveforms before the backbone has learned reliable self-similarity. Additionally, updating the entire model may intensify the learning burden of the backbone in the early training stage. 
Therefore, we adopt a decoupling strategy inspired by linear probing \cite{MoCo, MAE, DINO}, which detaches the gradient between the predictor and the backbone, as illustrated in Fig. \ref{SSPD}.
This strategy only update the predictor module for rPPG signal prediction, yielding a better performance experimentally. 
We will further discuss the rPPG signal decoupling strategy in Sec. \ref{ablation section}.

\begin{figure}
    \centering
    \includegraphics[height=4.6cm]{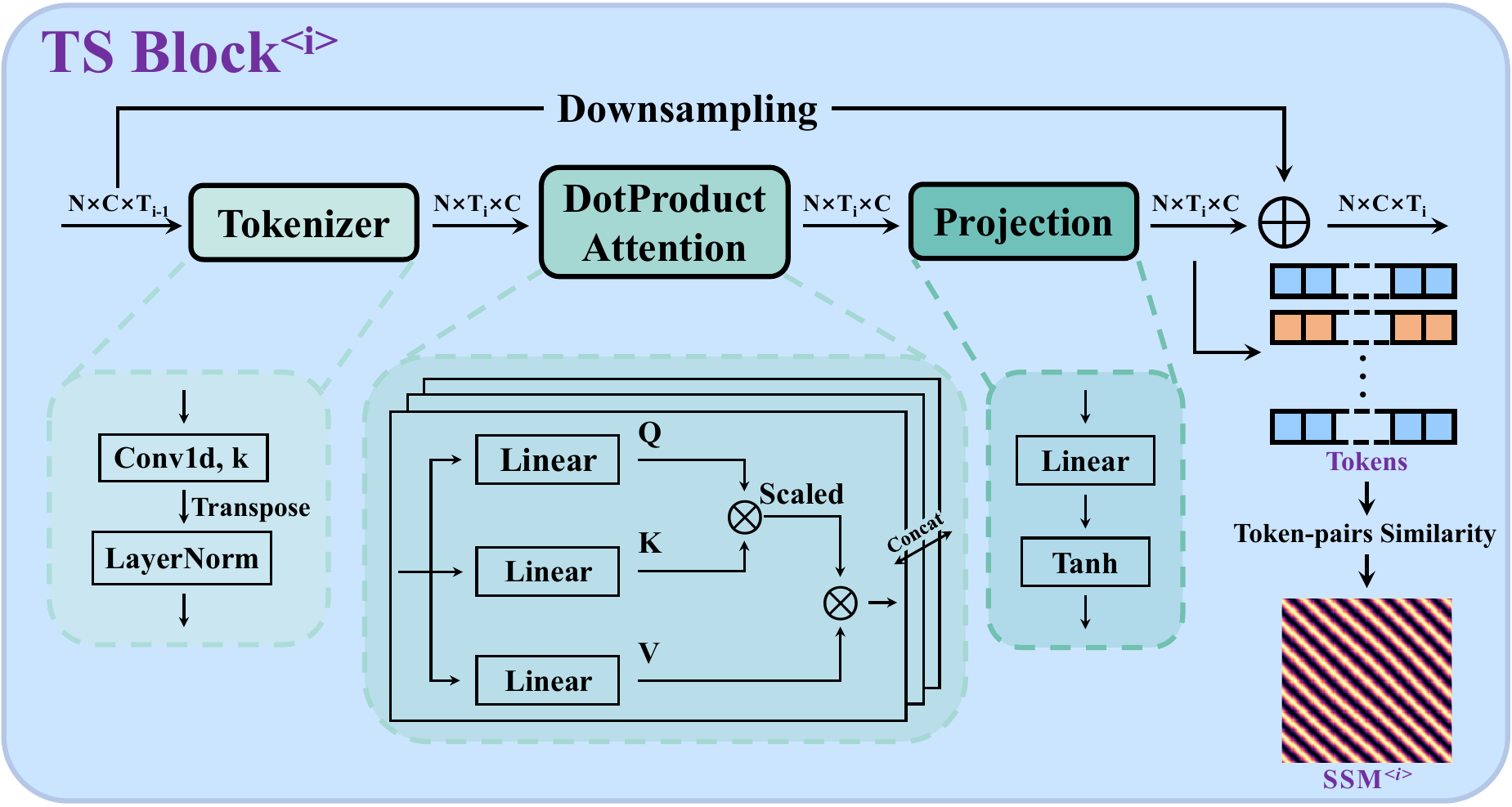}
    \caption{The architecture of the Temporal Similarity (TS) block. Firstly, the input sequences are projected into token embeddings at a specific time scale. Next, the multi-head dot-product attention improves the global context information, followed by a linear projection. We derive the self-similarity map which forms a layer in the temporal similarity pyramid from the token embeddings. We add a residual connection from the input sequence to the output tokens and perform downsampling for dimension alignment.}
    \label{TS block}
\end{figure}

\begin{figure}
    \centering
    \includegraphics[height=5.6cm]{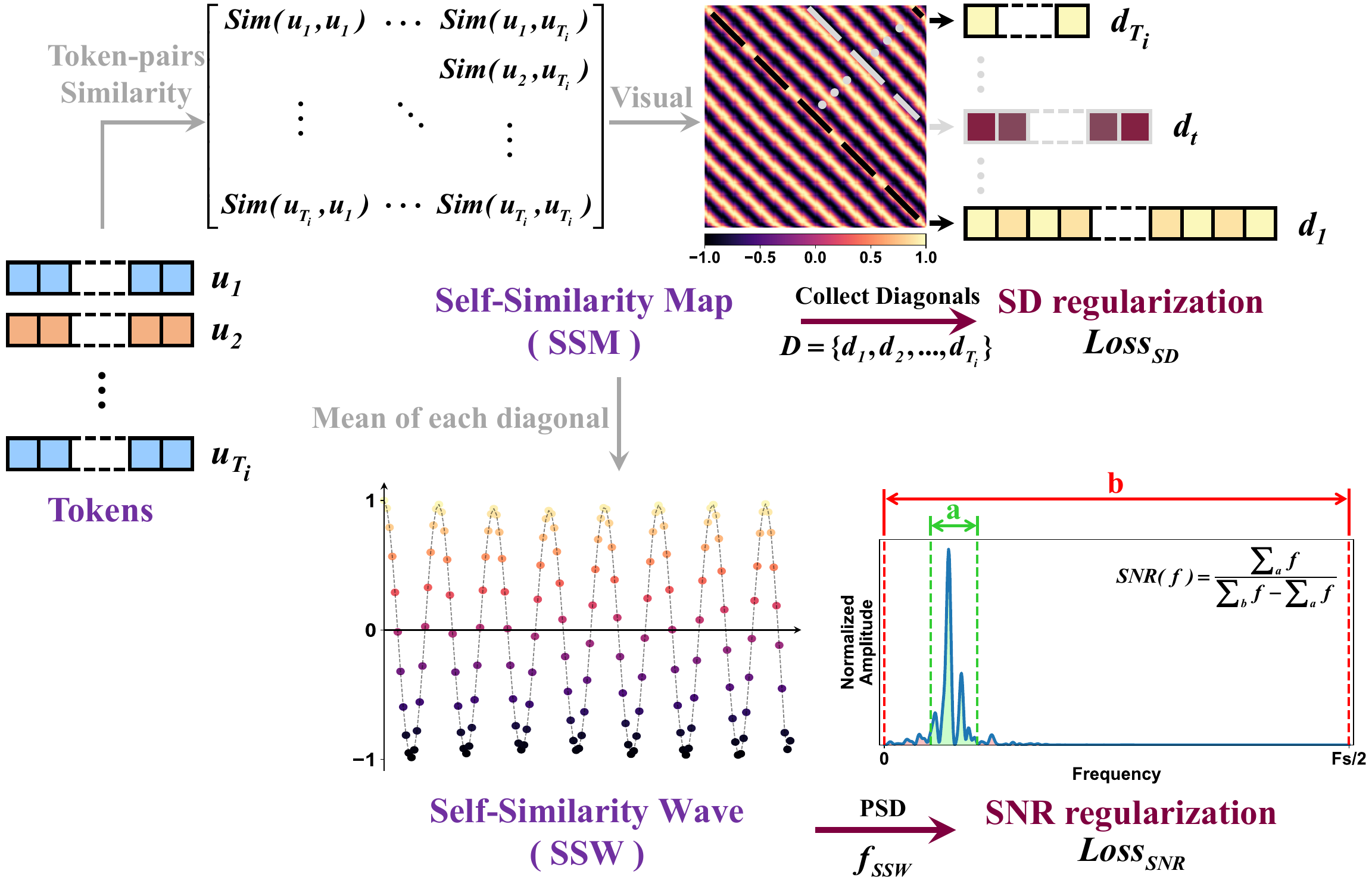}
    \caption{The derivation of self-similarity map, self-similarity wave, and periodicity regularizations. The self-similarity map is obtained by computing the cosine similarity between every pair of tokens. Subsequently, we calculate the mean value of each diagonal, yielding the self-similarity wave. The periodicity regularizations comprise the Standard Deviation (SD) regularization and Signal-to-Noise Ratio (SNR) regularization. SD regularization constrains the standard deviation of each diagonal from the self-similarity map, while SNR regularization constrains the SNR of the self-similarity wave.}
    \label{Self-similarity}
\end{figure}

\subsubsection{Separable Self-similarity Model}
To enable self-similarity-aware learning, we elaborate a Separable Self-Similarity Model ($S^3M$) that forces the network to learn multi-scale and long-distance physiological features better. Importantly, this model is separable and serves as a training auxiliary without any extra computing expenses in the inference stage. Specifically, the proposed $S^3M$ is composed of multiple Temporal Similarity (TS) blocks that transform the input sequences (i.e., $S_\mathcal{O}$, $S_\mathcal{T}$) into self-similarity maps,
which are subsequently incorporated into the temporal similarity pyramid. Each TS block within the model corresponds to a specific time scale, enabling the model to focus on multi-scale self-similar features that are synchronized with heartbeat events. In particular, as illustrated in Fig. \ref{TS block},
we initially tokenize the input sequence $S^{<i>}\in\mathbb{R}^{T_{i-1}\times C}$ into token embeddings $E^{<i>} = \{e_1^{<i>},e_2^{<i>},...,e_{T_{i}}^{<i>}\}$, where $e_t^{<i>}\in\mathbb{R}^{C}$ and $i$ denotes the i-th level TS block. Subsequently, we apply the multi-head dot-product attention \cite{Nonlocal} to enhance the global contextual relationships among the tokens, rather than utilizing self-attention directly. This is because the attention scores in self-attention are constrained within the range of (0, 1) due to the Softmax(·) normalization.
To better facilitate the learning of long-distance physiological features, we expand the attention score range to maintain symmetry between positive and negative values.
Specifically, we firstly project the input tokens $E^{<i>}$ to $K^{<i>}\in\mathbb{R}^{T_{i}\times C}$, $Q^{<i>}\in\mathbb{R}^{T_{i}\times C}$, $V^{<i>}\in\mathbb{R}^{T_{i}\times C}$, respectively.
Then, we calculate the attention output as shown below:

\begin{equation}
    \begin{aligned}
        A^{<i>} &= \frac{Q^{<i>}{K^{<i>}}^\mathsf{T}}{C}V^{<i>}
    \end{aligned}
    \label{DotProduct Attention}
\end{equation}

\noindent here, $\mathsf{T}$ stands for matrix transpose. Noteworthy, we also leverage multi-head joint modeling, which is the same as \cite{Transformer}. After that, we adopt another linear projection layer with parameters
$W^{<i>}\in\mathbb{R}^{C\times C}$ to $A^{<i>}$, resulting in $U^{<i>} = \{u_1^{<i>},u_2^{<i>},...,u_{T_{i}}^{<i>}\}$, where $u_t^{<i>}\in\mathbb{R}^{C}$. Next, we produce the self-similarity map as the final output of the TS block, as shown in Fig. \ref{TS block}. The self-similarity map
$\mathcal{M}^{<i>}\in\mathbb{R}^{T_{i}\times T_{i}}$ is formed by computing the similarity between each pair of tokens, as demonstrated in Eqn. \ref{SSM}:

\begin{equation}
    \begin{aligned}
        \mathcal{M}_{jk}^{<i>} = Sim(u_j^{<i>}, u_k^{<i>}) = \frac{u_j^{<i>} \cdot u_k^{<i>}}{\Vert u_j^{<i>} \Vert_2 \cdot \Vert u_k^{<i>} \Vert_2}
    \end{aligned}
    \label{Cosine similarity}
\end{equation}

Furthermore, we obtain the self-similarity wave $\mathcal{W}^{<i>}\in\mathbb{R}^{T_{i}}$ by calculating the mean value of each diagonal located in the upper triangle of the $\mathcal{M}^{<i>}$ matrix, as illustrated in Fig. \ref{Self-similarity}.
Specifically, we first retrieve the diagonals on the upper triangle of $\mathcal{M}^{<i>}$ to generate the diagonal set $D^{<i>} = \{d_1^{<i>},d_2^{<i>},...,d_{T_i}^{<i>}\}$, where $d_t^{<i>} = \{Sim(u_j^{<i>}, u_k^{<i>}) | j-k = t-1\}$, $t \in \left[1,T_{i}\right]$. Subsequently, we form the self-similarity wave in the same way as Eqn. \ref{SSW1}:

\begin{equation}
    \begin{aligned}
        \mathcal{W}^{<i>}_t = \frac{1}{T_i+1-t}\sum_{n=1}^{T_i+1-t} d_{t,n}^{<i>}
    \end{aligned}
    \label{SSW}
\end{equation}

In addition, we add a residual connection \cite{ResNet} from the input sequence $S^{<i>}$ to the output token embeddings $U^{<i>}$, and we utilize adaptive average pooling to maintain the alignment of the temporal dimension, which is illustrated below:

\begin{equation}
    \begin{aligned}
        S^{<i+1>} &= POOL_{T_{i-1}\rightarrow T_{i}}(S^{<i>}) + U^{<i>}
    \end{aligned}
    \label{Residual connection}
\end{equation}

The residual connection allows us to increase the number of TS blocks in $S^3M$ without significant degradation, facilitating the capture of self-similar representations under different time scales. Finally, we integrate the output of the TS blocks at multiple time scales, forming the temporal similarity pyramid
$\{\mathcal{M}^{<1>},\mathcal{M}^{<2>},...,\mathcal{M}^{<L>}\}$, along with the $\{\mathcal{W}^{<1>},\mathcal{W}^{<2>},...,\mathcal{W}^{<L>}\}$, where $L$ denotes the total number of blocks, i.e., the number of pyramid layers. 

The $S^3M$ contributes significantly to extracting more reliable physiological features through self-similarity-aware learning, we further make it separable to guarantee that there are no extra computational expenses in the inference stage due to the incorporation of $S^3M$. To this end, we devise two distinct feature flow paths for training and inference, as depicted in Fig. \ref{SSPD}.
During the training stage, feature maps are propagated through all modules in the SSPD framework, resulting in the rPPG prediction
$y_\mathcal{O}$, $y_\mathcal{T}$, and the temporal similarity pyramid $\{\mathcal{M}_\mathcal{O}^{<1>},\mathcal{M}_\mathcal{O}^{<2>},...,\mathcal{M}_\mathcal{O}^{<L>}\}$, $\{\mathcal{M}_\mathcal{T}^{<1>},\mathcal{M}_\mathcal{T}^{<2>},...,\mathcal{M}_\mathcal{T}^{<L>}\}$ for self-distillation. While in the inference stage, all TS blocks in $S^3M$ are discarded, and the rPPG estimation is obtained solely from the backbone and subsequent predictor module. 
Consequently, by using the $S^3M$, our framework can facilitate better learning of multi-scale and long-distance physiological features without increasing the computational cost during inference. 

\begin{algorithm}
	\caption{SSPD PyTorch-like Pseudocode}
    \label{Pseudocode}
	\KwIn{\\
    $\theta_\mathcal{O}$, $\theta_\mathcal{T}$;   \hspace{55pt}   \textcolor{BLUE}{\# online and target models} \\
    $x$;            \hspace{78.6pt}   \textcolor{BLUE}{\# input clip} \\
    LGA;            \hspace{63.7pt}   \textcolor{BLUE}{\# local-global augmentation} \\
    MDM;            \hspace{59.3pt}   \textcolor{BLUE}{\# masked difference modeling} \\
    $\mathcal{L}_{Pearson}$;      \hspace{46pt}   \textcolor{BLUE}{\# negative Pearson loss} \\
    $\mathcal{L}_{SD}$, $\mathcal{L}_{SNR}$;      \hspace{33.1pt}   \textcolor{BLUE}{\# SD and SNR regularizations} \\
    }
    $\theta_\mathcal{O}$.params = $\theta_\mathcal{T}$.params \\
    \For{$x$ in loader}{
		    $v_l$, $v_g$ = LGA$_\mathcal{O}$($x$), LGA$_\mathcal{T}$($x$) \\
        $v_m$, $v_o$ = MDM$_\mathcal{O}$($v_l$), MDM$_\mathcal{T}$($v_g$) \\
        \textcolor{BLUE}{\# rPPG, temporal similarity pyramid} \\
        $y_\mathcal{O}$, $\mathcal{M}_\mathcal{O}$ = $\theta_\mathcal{O}$($v_m$)  \\
        $y_\mathcal{T}$, $\mathcal{M}_\mathcal{T}$ = $\theta_\mathcal{T}$($v_o$) \\
        
        $\mathcal{L}_{SSPD}$ = D($y_\mathcal{O}$, $\mathcal{M}_\mathcal{O}$, $y_\mathcal{T}$, $\mathcal{M}_\mathcal{T}$) + $\alpha \cdot \mathcal{L}_{SD}$($\mathcal{M}_\mathcal{O})$  \\
        \hspace{41pt} + $\beta \cdot \mathcal{L}_{SNR}$($y_\mathcal{O}, \mathcal{M}_\mathcal{O}$)
               
        $\mathcal{L}_{SSPD}$.backward() \\
        update($\theta_\mathcal{O}$) \\
        $\theta_\mathcal{T}$.params = $\rho$·$\theta_\mathcal{T}$.params + (1$-\rho$)·$\theta_\mathcal{O}$.params
	}

    \SetKwProg{Fn}{def}{:}{}
    \Fn {\rm D({$r_1$, $p_1$, $r_2$, $p_2$})}{
        $r_2$, $p_2$ = $r_2$.detach(), $p_2$.detach() \\
        \textcolor{BLUE}{\# PSD}  \\
        $f_1$, $f_2$ = FFT($r_1$), FFT($r_2$) \hspace{18.5pt}  \\  
        \textcolor{BLUE}{\# SSM $\rightarrow$ SSW}  \\
        $w_1$, $w_2$ = Diagonals($p_1$), Diagonals($p_2$) \hspace{6.5pt}  \\
        $\mathcal{L}_{Distill}$ = $\mathcal{L}_{Pearson}$($r_1$, $r_2$) + $\mathcal{L}_{MSE}$($f_1$, $f_2$) + \\
        \hspace{40pt} $\mathcal{L}_{MSE}$($p_1$, $p_2$) + $\mathcal{L}_{MSE}$($w_1$, $w_2$) \\
        \KwRet $\mathcal{L}_{Distill}$
    }
\end{algorithm}

\subsection{Hierarchical Self-distillation}
Our SSPD adopts the same overall structure as the latest self-distillation frameworks \cite{DINO, iBOT}. 
On the one hand, the online model is supervised by the output of the target model and updates its parameters via gradient descent. 
On the other hand, the target model detaches all gradients to prevent model collapse \cite{SimSiam} (Since there are no negative samples), 
and its parameters are updated by the online model through Exponential Moving Average (EMA) \cite{DINO, iBOT}. 
Under this paradigm, the online and target models share the same structure, 
and the target model can be seen as an ensemble of the past iterations of the online model. 
Moreover, while the online model has a higher noise level input $V_m$, 
the target model's input $V_o$ is more informative, progressively providing higher-confidence supervision to the online model as training progresses. 
However, since SSPD performs both self-similarity-aware learning and rPPG signal decoupling, 
we propose a hierarchical distillation strategy that includes Temporal Similarity Pyramid Distillation (TSPD) and RPPG Prediction Distillation (RPD). 
As mentioned in Sec. \ref{Self-similarity Prior in rPPG}, 
directly distilling fine-grained SSM shared by the online and target models is challenging. Thus,
we introduce two regularizations based on prior knowledge illustrated in Fig. \ref{SSM_SSW} to learn the strong periodic components in SSM through SSW. 
These regularizations enhance SSPD's ability to capture reliable self-similar physiological features, particularly in the early stages of training.

\subsubsection{Temporal Similarity Pyramid Distillation}
First, we propose Temporal Similarity Pyramid Distillation (TSPD) to enable self-similarity-aware learning.
Specifically, the temporal similarity pyramids (i.e., $\{\mathcal{M}_\mathcal{O}^{<1>},\mathcal{M}_\mathcal{O}^{<2>},...,\mathcal{M}_\mathcal{O}^{<L>}\}$, $\{\mathcal{M}_\mathcal{T}^{<1>},\mathcal{M}_\mathcal{T}^{<2>},...,\mathcal{M}_\mathcal{T}^{<L>}\}$) indicate the self-similar representations induced by heartbeat rhythms that are shared between the online and target models. 
We distill SSM and SSW in each layer of the temporal similarity pyramid by TSPD loss defined as follows:

\begin{equation}
    \begin{aligned}
        \mathcal{L}_{TSPD} = \hspace{0.1cm}&\frac{1}{N}\sum_{i=1}^{N}\sum_{j=1}^{L}(\left\lVert{\mathcal{M}_\mathcal{O}^{<j>}}^i-{\mathcal{M}_\mathcal{T}^{<j>}}^i\right\rVert^2_2\\
        &+\left\lVert {\mathcal{W}_\mathcal{O}^{<j>}}^i-{\mathcal{W}_\mathcal{T}^{<j>}}^i\right\rVert^2_2) 
    \end{aligned}
    \label{TSPD loss}
\end{equation}
\noindent here N is the number of samples.

\begin{table*}[t]
    \centering
    \setlength{\abovecaptionskip}{8pt} 
    \caption{Intra-dataset and cross-dataset HR estimation results compared to the state-of-the-art methods. The best results are highlighted in \textbf{bold}. The notation ``-L3'' indicates that the temporal similarity pyramid consists of three layers, while ``$\dagger$'' denotes the model is trained on the VIPL-HR dataset and evaluated on other datasets.}
    \label{SSL result on small datasets}
    \resizebox{2         \columnwidth}{!}{
    \setlength{\tabcolsep}{2.7mm}{}
    \begin{tabular}{clccccccccc}
    \toprule[1pt]                &       & \multicolumn{3}{c}{UBFC-rPPG}       & \multicolumn{3}{c}{PURE}     & \multicolumn{3}{c}{MR-NIRP}     \\ \cmidrule(lr){3-5} \cmidrule(lr){6-8} \cmidrule(lr){9-11} 
    \multirow{-2}{*}{\begin{tabular}[c]{@{}c@{}}Method\\ Types\end{tabular}} & \multirow{-2}{*}{Methods}     & \begin{tabular}[c]{@{}c@{}}MAE$\downarrow$\end{tabular} & \begin{tabular}[c]{@{}c@{}}RMSE$\downarrow$\end{tabular} & R$\uparrow$    & \begin{tabular}[c]{@{}c@{}}MAE$\downarrow$ \end{tabular} & \begin{tabular}[c]{@{}c@{}}RMSE$\downarrow$ \end{tabular} & R$\uparrow$        & \begin{tabular}[c]{@{}c@{}}MAE$\downarrow$ \end{tabular} & \begin{tabular}[c]{@{}c@{}}RMSE$\downarrow$ \end{tabular} & R$\uparrow$   \\ \midrule
                                 & GREEN \cite{GREEN}                     & 8.33              & 10.88             & 0.48              & 5.83           & 10.64         & 0.59              & -                 & -              & -            \\
                                 & POS \cite{POS}                         & 7.80              & 11.44             & 0.62              & 2.99           & 4.79          & 0.94              & -                 & -              & -            \\ 
     \multirow{-2}{*}{Traditional}  & CHROM \cite{CHROM}                  & 6.69              & 8.82              & 0.82              & 4.17           & 6.26          & 0.92              & -                 & -              & -            \\
                                 & ICA \cite{ICA}                         & 5.63              & 8.53              & 0.71              & 2.59           & 4.23          & 0.94              & -                 & -              & -            \\ \midrule
                                 & CAN \cite{CAN}                         & -                 & -                 & -                 & 1.27           & 3.06          & 0.97              & 7.78              & 16.8           & -0.03        \\
                                 & HR-CNN \cite{HR-CNN}                   & -                 & -                 & -                 & 1.84           & 2.37          & 0.98              & -                 & -              & -            \\
                                 & SynRhythm \cite{SynRhythm}             & 5.59              & 6.82              & 0.72              & -              & -             & -                 & -                 & -              & -            \\
                                 & PhysNet \cite{PhysNet}                 & -                 & -                 & -                 & 2.1            & 2.6           & \textbf{0.99}     & 3.07              & 7.55           & 0.655        \\
                                 & PulseGAN \cite{PulseGAN}               & 1.19              & 2.10              & 0.98              & -              & -             & -                 & -                 & -              & -            \\
                                 & Dual-GAN \cite{Dual-GAN}               & \textbf{0.44}     & \textbf{0.67}     & \textbf{0.99}     & 0.82           & 1.31          & \textbf{0.99}     & -                 & -              & -            \\
     \multirow{-7}{*}{Supervised}& Nowara2021 \cite{Nowara2021}           & -                 & -                 & -                 & -              & -             & -                 & \textbf{2.34}     & 4.46           & 0.85         \\ \midrule
                                 & RemotePPG \cite{RemotePPG}             & 1.85              & 4.28              & 0.93              & 2.3            & 2.9           & \textbf{0.99}     & 4.75              & 9.14           & 0.61         \\
                                 & Contrast-Phys \cite{Contrast-Phys}     & 0.64              & 1.00              & \textbf{0.99}     & 1.00           & 1.40          & \textbf{0.99}     & 2.68              & 4.77           & 0.85         \\
     \multirow{3}{*}{Unsupervised}  & \cellcolor[HTML]{FFE5E5}SSPD-L3 & \cellcolor[HTML]{FFE5E5}0.50   & \cellcolor[HTML]{FFE5E5}0.71  & \cellcolor[HTML]{FFE5E5}\textbf{0.99}      & \cellcolor[HTML]{FFE5E5}\textbf{0.53}      & \cellcolor[HTML]{FFE5E5}\textbf{1.16}   & \cellcolor[HTML]{FFE5E5}\textbf{0.99} & \cellcolor[HTML]{FFE5E5}2.65      & \cellcolor[HTML]{FFE5E5}\textbf{3.66}        & \cellcolor[HTML]{FFE5E5}\textbf{0.94} \\ \cline{2-11}
                                 & \emph{cross-dataset evaluations} \\
                                 & RemotePPG$^\dagger$ \cite{RemotePPG}          & 4.67   & 8.75  & 0.72      & 2.22   & 3.82     & 0.96     & 7.93     & 11.01    & 0.05 \\                                                 
                                 & Contrast-Phys$^\dagger$ \cite{Contrast-Phys}  & 1.90   & 3.30  & 0.95      & 3.07   & 4.85     & 0.94     & 5.73     & 9.18     & 0.18 \\                                                      
                                 & \cellcolor[HTML]{FFE5E5}SSPD-L3$^\dagger$  & \cellcolor[HTML]{FFE5E5}1.08   & \cellcolor[HTML]{FFE5E5}1.81  & \cellcolor[HTML]{FFE5E5}\textbf{0.99}      & \cellcolor[HTML]{FFE5E5}1.24      & \cellcolor[HTML]{FFE5E5}1.77     & \cellcolor[HTML]{FFE5E5}\textbf{0.99} & \cellcolor[HTML]{FFE5E5}3.52      & \cellcolor[HTML]{FFE5E5}6.21    & \cellcolor[HTML]{FFE5E5}0.73 \\ \bottomrule[1pt]                                                   
    \end{tabular}
    }
\end{table*}

\subsubsection{RPPG Prediction Distillation}
After self-similar representation learning, we propose RPPG Prediction Distillation (RPD) for rPPG signal decoupling.
RPD distills the shared rPPG output (i.e., $y_\mathcal{O}$, $y_\mathcal{T}$) between two distorted input views. 
We calculate the normalized PSD noted as $f_{y_\mathcal{O}}$, $f_{y_\mathcal{T}}$, and the RPD loss is described as follows:

\begin{equation}
    \begin{aligned}
        \mathcal{L}_{RPD} &= \mathcal{L}_{Pearson}(y_\mathcal{O}, y_\mathcal{T}) + \frac{1}{N}\sum_{i=1}^{N}\left\lVert f_{y_{\mathcal{O}}}^i-f_{y_{\mathcal{T}}}^i \right\rVert^2_2
    \end{aligned}
    \label{RPD loss}
\end{equation}
here $\mathcal{L}_{Pearson}(\cdot)$ means negative Pearson loss \cite{PhysNet}.

Finally, our hierarchical self-distillation loss is the sum of TSPD loss and RPD loss:

\begin{equation}
    \begin{aligned}
        \mathcal{L}_{Distill} = \mathcal{L}_{TSPD} + \mathcal{L}_{RPD}
    \end{aligned}
    \label{Distill loss}
\end{equation}

\subsubsection{Periodicity Regularizations}
As discussed in Sec. \ref{Self-similarity Prior in rPPG}, the smooth heart rate prior allows us to extract the periodicity in SSM by constructing SSW, 
which helps SSPD learn more reliable periodic physiological features. To emphasize the periodicity in self-similarity-aware learning, we propose Standard Deviation (SD) and Signal-to-Noise Ratio (SNR) regularizations based on the two prior knowledge in Fig. \ref{SSM_SSW}, respectively.

First, we propose the SD regularization, which capitalizes on the prior knowledge that \emph{different tokens are repeated in a specific time interval}.
As depicted in Fig. \ref{Self-similarity}, we retrieve the diagonals on the upper triangle of $\mathcal{M}^{<i>}$ to form the diagonal set
$D^{<i>} = \{d_1^{<i>},d_2^{<i>},...,d_{T_i}^{<i>}\}$, where $d_t^{<i>}\in\mathbb{R}^{T_i+1-t}$. Then we compute the SD regularization loss as follows:

\begin{equation}
    \begin{aligned}
        \mathcal{L}_{SD} = \frac{1}{N}\frac{1}{L}\frac{1}{T_j}\sum_{i=1}^{N}\sum_{j=1}^{L}\sum_{k=1}^{T_j}SD(\varepsilon \cdot (T_j+1-k)\cdot{d_k^{<j>}}^i)
    \end{aligned}
    \label{SD loss}
\end{equation}
where the $SD(\cdot)$ means standard deviation \cite{EVM-CNN}, and we use $\varepsilon$ to modulate the range of the SD output. Experimentally, $\varepsilon$ is 0.05 by default.

Moreover, since \emph{tokens of different time intervals are repeated periodically}, we propose SNR regularization to constrain periodicity shared in SSM, SSW, and the predicted rPPG within the prior frequency band of heart rate.
As shown in Fig. \ref{Self-similarity}, we calculate the normalized PSD of the $\mathcal{W}^{<i>}$ denoted as $f_{\mathcal{W}^{<i>}}$ and we have obtained $f_{y_{\mathcal{O}}}$ in Eqn. \ref{RPD loss}. 
Then, we adjust the definition of SNR as proposed in \cite{SparsePPG}:

\begin{equation}
    \begin{aligned}
        SNR(f) = \frac{\sum_af}{\sum_bf-\sum_af}
    \end{aligned}
    \label{SNR}
\end{equation}

\noindent In our work, $a$ represents the frequency range from 0.65 Hz to 3 Hz, which corresponds to the typical range of heart rate frequencies \cite{RemotePPG}.
$b$ includes the entire frequency band, ranging from 0 Hz to $F_s/2$ Hz, where $F_s$ is the sampling rate. Given these definitions, we propose the SNR regularization as follows:

\begin{equation}
    \begin{aligned}
        \mathcal{L}_{SNR} = \frac{1}{N}\frac{1}{L}\sum_{i=1}^{N}(\sum_{j=1}^{L}\frac{1}{SNR({f_{\mathcal{W}^{<j>}}}^i)} + \frac{1}{SNR({f_{y_{\mathcal{O}}}}^i)})
    \end{aligned}
    \label{SNR loss}
\end{equation}

Experimentally, these two periodicity regularizations boost more performance in the more challenging dataset.
To sum up, the total loss of the SSPD framework is summarized as follows:

\begin{equation}
    \begin{aligned}
        \mathcal{L}_{SSPD} = \mathcal{L}_{Distill} + \alpha \cdot \mathcal{L}_{SD} + \beta \cdot \mathcal{L}_{SNR}
    \end{aligned}
    \label{total loss}
\end{equation}
where hyperparameters $\alpha$, $\beta$ equal to 0.8, 0.6, respectively. During training, we only optimize the online model by back-propagation, while the parameters of the target model are directly updated using EMA:

\begin{equation}
    \begin{aligned}
        \theta_\mathcal{T} \leftarrow \rho \cdot \theta_\mathcal{T} + (1-\rho) \cdot \theta_\mathcal{O}
    \end{aligned}
    \label{EMA}
\end{equation}
here $\theta_\mathcal{O}$ and $\theta_\mathcal{T}$ represent the parameters of the online and target models, respectively. The momentum rate, denoted by $\rho$, is typically set to 0.9. We propose a pseudo-code in Algorithm \ref{Pseudocode} to show the overall implementation of our SSPD framework.

\section{Experiments}
\subsection{Datasets and Metrics}
We conduct experiments on four commonly used open-access benchmarks: PURE \cite{PURE}, UBFC-rPPG \cite{UBFC}, VIPL-HR \cite{VIPL}, and MR-NIRP \cite{NIRP}. Specifically, \textbf{PURE} \cite{PURE} consists of 10 persons with 6 different head motion types. We adopt the pre-defined strategy that aligns with previous works \cite{Contrast-Phys, HR-CNN} to split the training and test sets. \textbf{UBFC-rPPG} \cite{UBFC} comprises 42 subjects who are asked to play a mathematical game to elicit non-stationary HRs. We follow the protocol established by the dataset author to split the training and test sets, which is consistent with \cite{Dual-GAN, Contrast-Phys}. \textbf{VIPL-HR} \cite{VIPL} is a challenging large-scale benchmark for remote physiological measurement, it contains 9 scenarios for 107 subjects with different head movements, illumination changes, and recording devices. We use a subject-exclusive 5-fold cross-validation protocol in line with the \cite{Dual-GAN, PhysFormer}. \textbf{MR-NIRP} \cite{NIRP} is another challenging dataset due to the low intensity of the rPPG signals in NIR videos. This dataset comprises 8 subjects and 2 experiments (still and motion), and we implement the subject-exclusive 8-fold cross-validation strategy as utilized in \cite{Contrast-Phys}.

We apply the widely used evaluation metrics in remote HR measurement \cite{EVM-CNN}, including Mean Absolute Error (MAE), Root Mean Square Error (RMSE), Standard Deviation (SD), and Pearson's correlation coefficient (R). 
Noteworthy, these four metrics evaluate the performance of the physiological measurement model from distinct perspectives. First, MAE/RMSE emphasizes the absolute error between the predicted heart rate and the ground truth, assessing the model's accuracy. Besides, SD centers on error stability, providing insights into the generalization performance. Finally, R underscores the linear relationship between the overall distribution of the predicted heart rate and the ground truth.

\subsection{Implementation Details}
Our framework is implemented on NVIDIA GeForce RTX 3090 GPU using Pytorch 2.0.0. We train the model for 50 epochs via Adam optimizer \cite{adam} with no weight decay, and the learning rate is 0.001. We set the initial and ending momentum rate to 0.9 and 1.0, respectively, which is scheduled by cosine annealing \cite{CosineAnnealing}. In the training stage, 10s video clips with a size of $128 \times 128$ are selected randomly from the input video, with a batch size of 8. During inference, we strictly follow the previous works \cite{PhysFormer, Contrast-Phys} and the rPPG signal is typically predicted from each 30s non-overlapping clip in the test set unless otherwise indicated. We detect all the peaks in the estimated rPPG waveform, and the HR result is obtained based on the averaged peak intervals.



\begin{table}
    \vspace{-0.2cm}
    \centering
    \caption{HR estimation results of 5-fold cross-validation on the VIPL-HR dataset. The best results are highlighted in \textbf{bold}. The $^\blacklozenge$ denotes that the evaluation protocol aligns with the \cite{ViViT}, except that SSPD does not require any labels for fine-tuning.}
    \label{SSL VIPL result}
    \resizebox{1.0                                                             \columnwidth}{!}{
    \setlength{\tabcolsep}{1.2mm}{}
    \begin{tabular}{clcccc}
    \toprule
    \begin{tabular}[c]{@{}c@{}}Method\\ Types\end{tabular} & \multicolumn{1}{l}{Methods} & \begin{tabular}[c]{@{}c@{}}SD$\downarrow$ \\ \end{tabular} & \begin{tabular}[c]{@{}c@{}}MAE$\downarrow$ \\ \end{tabular} & \begin{tabular}[c]{@{}c@{}}RMSE$\downarrow$ \\ \end{tabular}  & R$\uparrow$ \\ \midrule
    \multirow{4}{*}{Traditional}    & GREEN \cite{GREEN}                      & 8.93    & 11.34    & 14.41     & 0.16  \\
                                    & POS \cite{POS}                          & 11.57   & 15.89    & 19.63     & 0.52  \\
                                    & CHROM \cite{CHROM}                      & 11.39   & 13.88    & 17.93     & 0.45  \\ 
                                    & ICA \cite{ICA}                          & 9.75    & 10.68    & 14.44     & 0.17  \\ \midrule
                                     
    \multirow{5}{*}{Supervised}     & PhysNet \cite{PhysNet}                  & 14.9   & 10.8    & 14.8   & 0.20   \\
                                    & DeepPhys \cite{Deepphys}                & 13.6   & 11.0    & 13.8   & 0.11   \\
                                    & RhythmNet \cite{RhythmNet}              & 8.11   & 5.30    & 8.14   & 0.76   \\
                                    & PhysFormer \cite{PhysFormer}            & 7.74   & 4.97    & 7.79   & 0.78   \\
                                    & Dual-GAN \cite{Dual-GAN}                & 7.63   & \textbf{4.93}    & \textbf{7.68}   & \textbf{0.81}   \\ \midrule
    \multirow{5}{*}{Unsupervised}   & ViViT$^\blacklozenge$ \cite{ViViT}      & 14.15  & 10.75   & 14.20  & -0.039 \\
                                    & \cellcolor[HTML]{FFE5E5}SSPD-L3-160$^\blacklozenge$  & \cellcolor[HTML]{FFE5E5}9.69   & \cellcolor[HTML]{FFE5E5}8.37  & \cellcolor[HTML]{FFE5E5}12.79  & \cellcolor[HTML]{FFE5E5}0.54  \\
                                    & RemotePPG \cite{RemotePPG}              & 10.85  & 8.50   & 13.76  & 0.31 \\
                                    & Contrast-Phys \cite{Contrast-Phys}      & 8.04  & 8.80   & 11.92  & 0.24 \\
                                    & \cellcolor[HTML]{FFE5E5}SSPD-L3         & \cellcolor[HTML]{FFE5E5}\textbf{7.42}      & \cellcolor[HTML]{FFE5E5}6.04      & \cellcolor[HTML]{FFE5E5}9.56       & \cellcolor[HTML]{FFE5E5}0.59  \\ \bottomrule
    \end{tabular}
    }
\end{table}

\begin{figure}
    \centering
    \includegraphics[height=4.8cm]{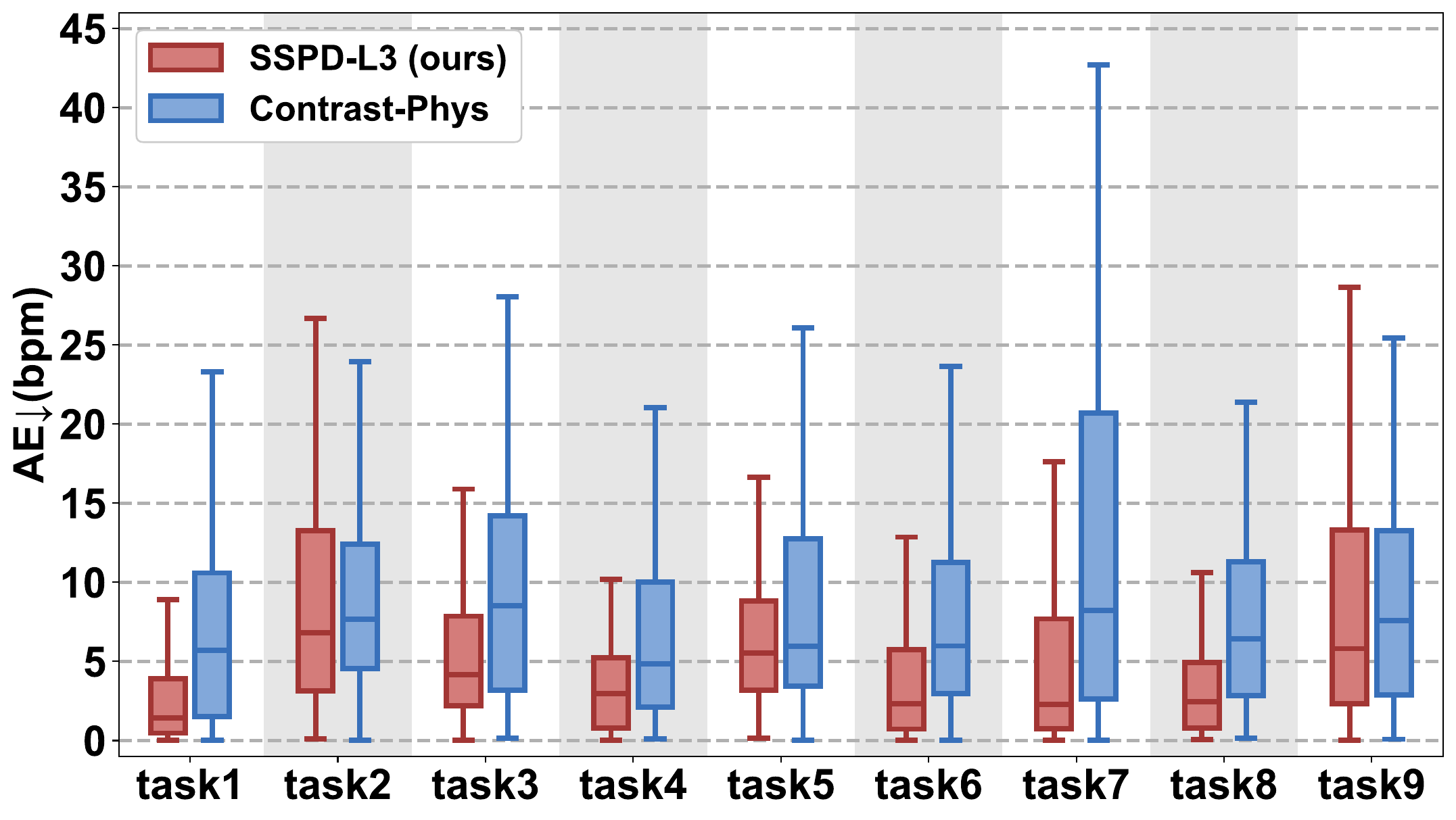}
    \caption{The distribution of absolute error (AE) across different scenarios for all subjects in the VIPL-HR dataset. AE is measured as the difference between the estimated HR and the ground truth HR. The 9 scenarios in the dataset are represented by task1-task9, which correspond to stable, motion, talking, bright, dark, long distance, exercise, phone stable, and phone motion scenarios, respectively.}
    \label{VIPL plot}
    \vspace{-10pt}
\end{figure}

\begin{figure*}[t]
    \centering
    \subfloat[SSPD (ours)]{\includegraphics[width=0.248\textwidth]{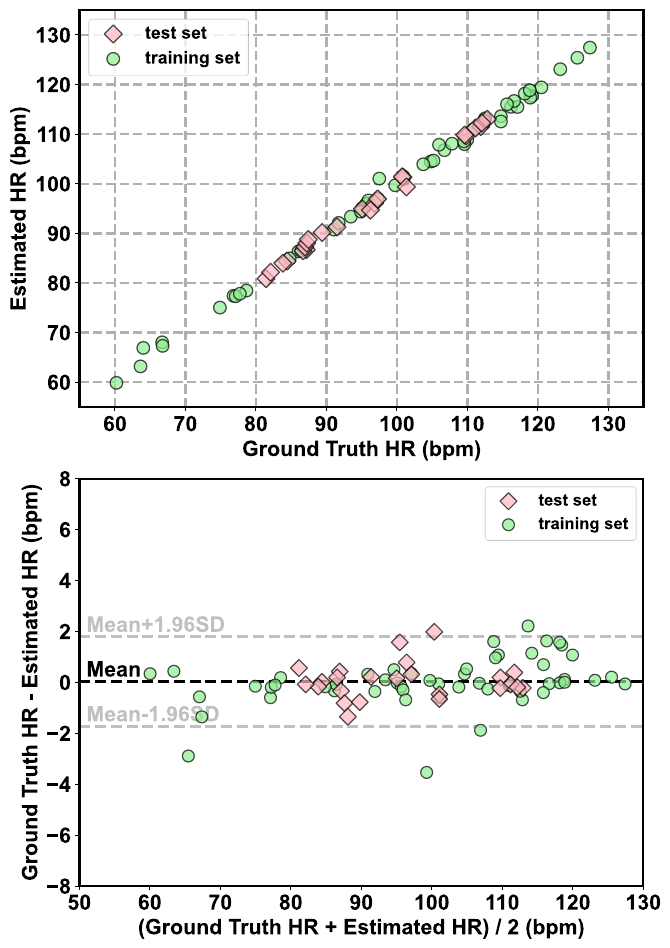}}
    \subfloat[Contrast-Phys \cite{Contrast-Phys}]{\includegraphics[width=0.248\textwidth]{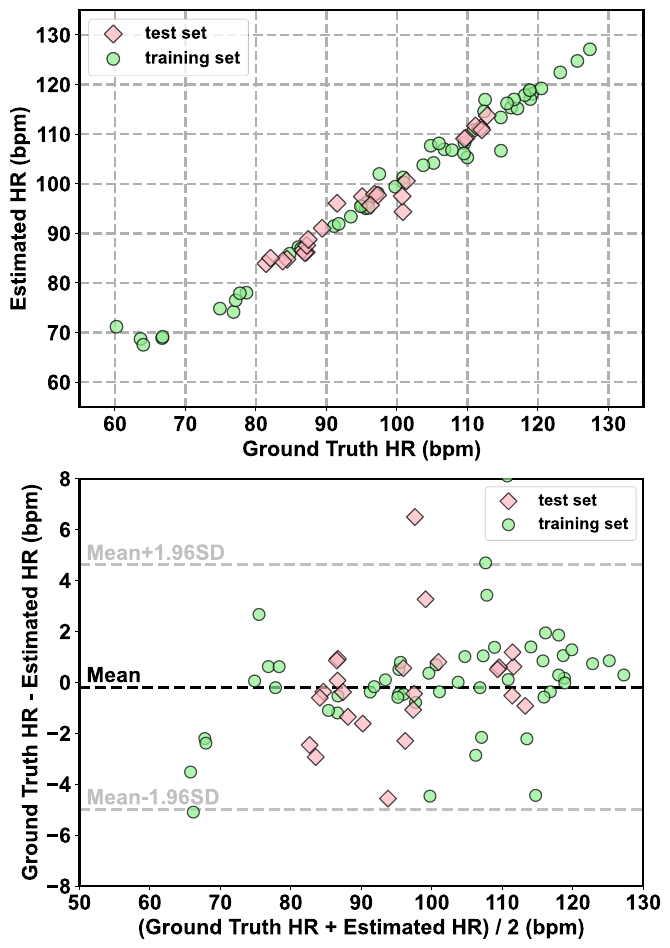}}
    \subfloat[RemotePPG \cite{RemotePPG}]{\includegraphics[width=0.248\textwidth]{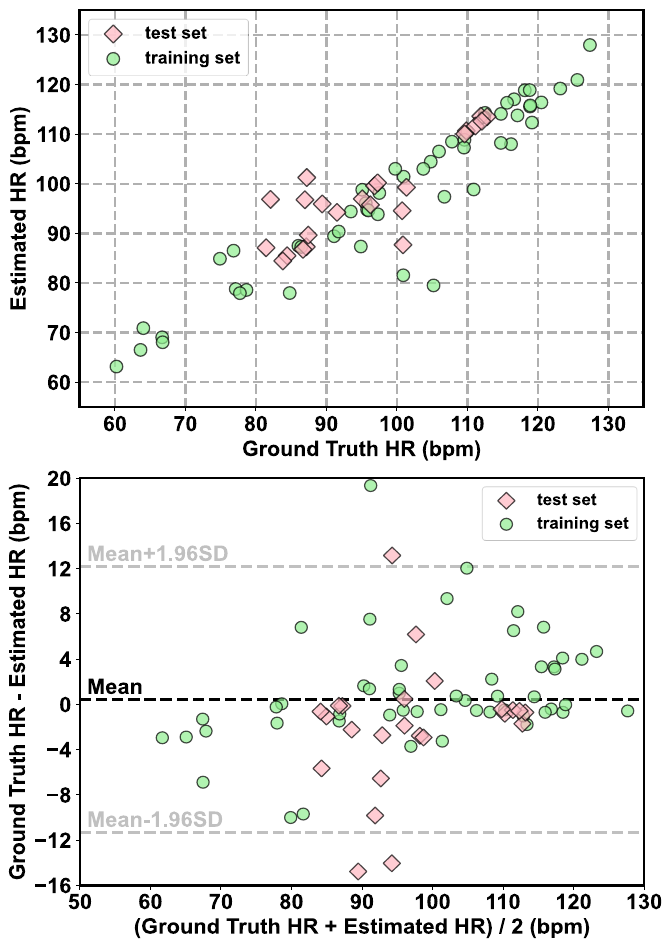}}
    \subfloat[ICA \cite{ICA}]{\includegraphics[width=0.248\textwidth]{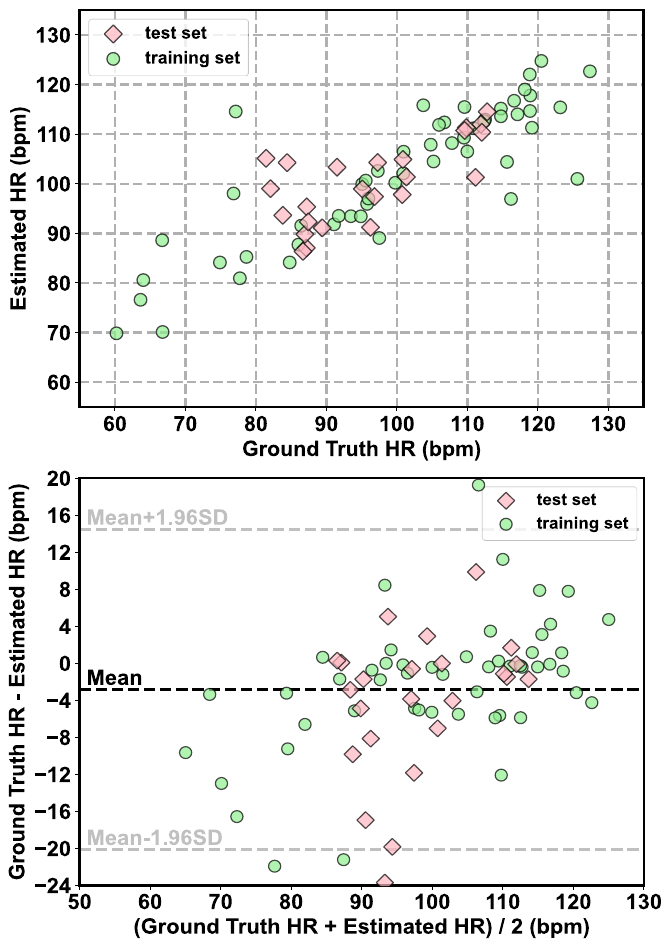}}
    \caption{The scatter plots (top) and Bland-Altman plots (bottom) illustrate the correlation between ground truth HR and estimated HR on the UBFC-rPPG dataset.} 
    \label{UBFC visualization}
 \end{figure*}

\begin{figure*}[t]
    \centering
    \subfloat[SSPD (ours)]{
        \includegraphics[width=0.325\textwidth]{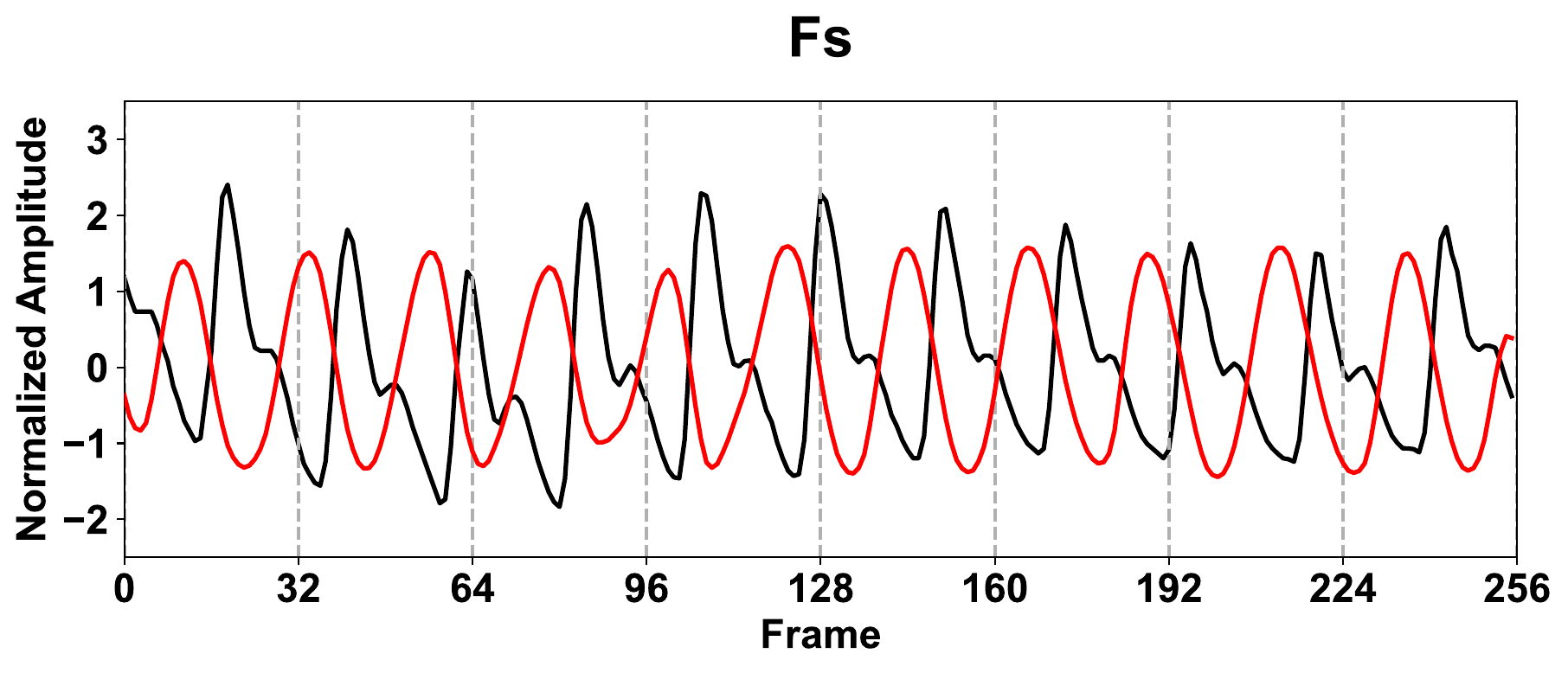}
        \includegraphics[width=0.325\textwidth]{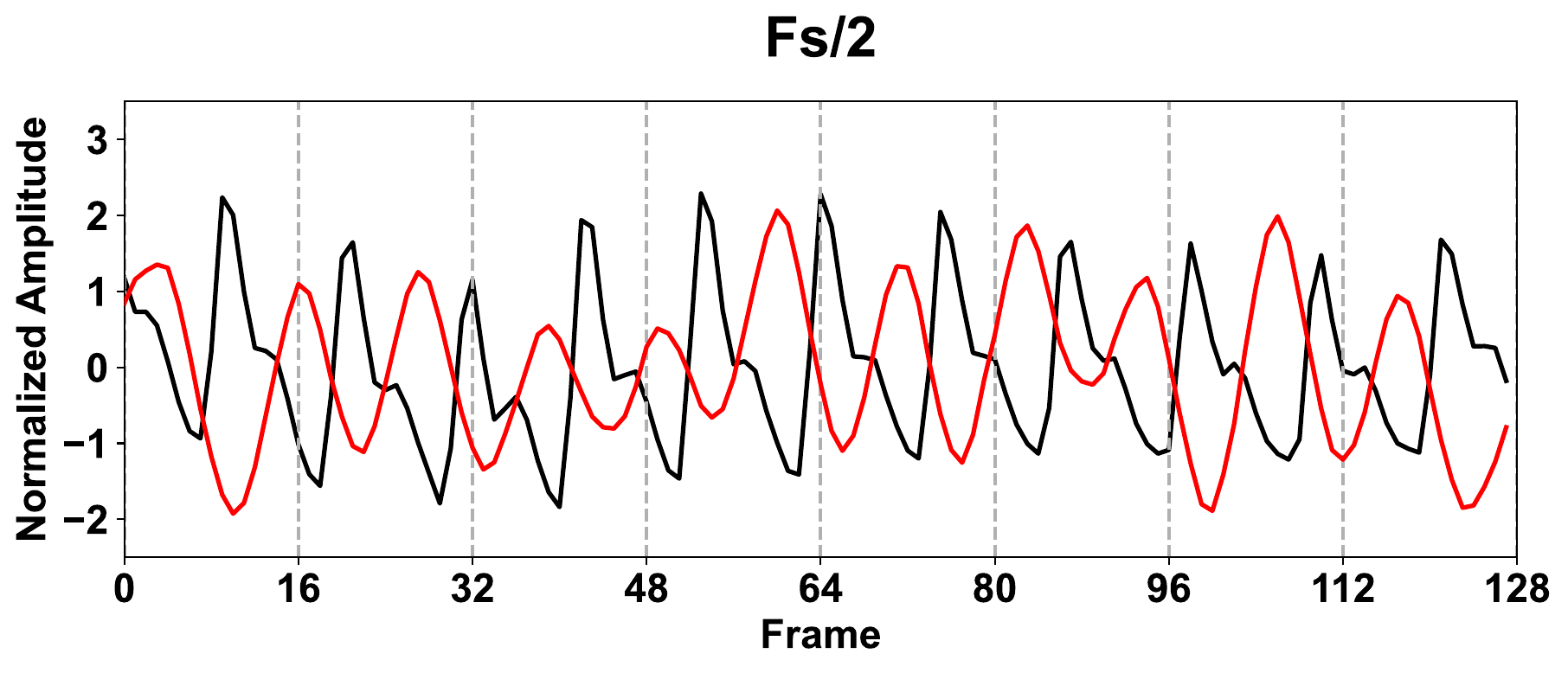}
        \includegraphics[width=0.325\textwidth]{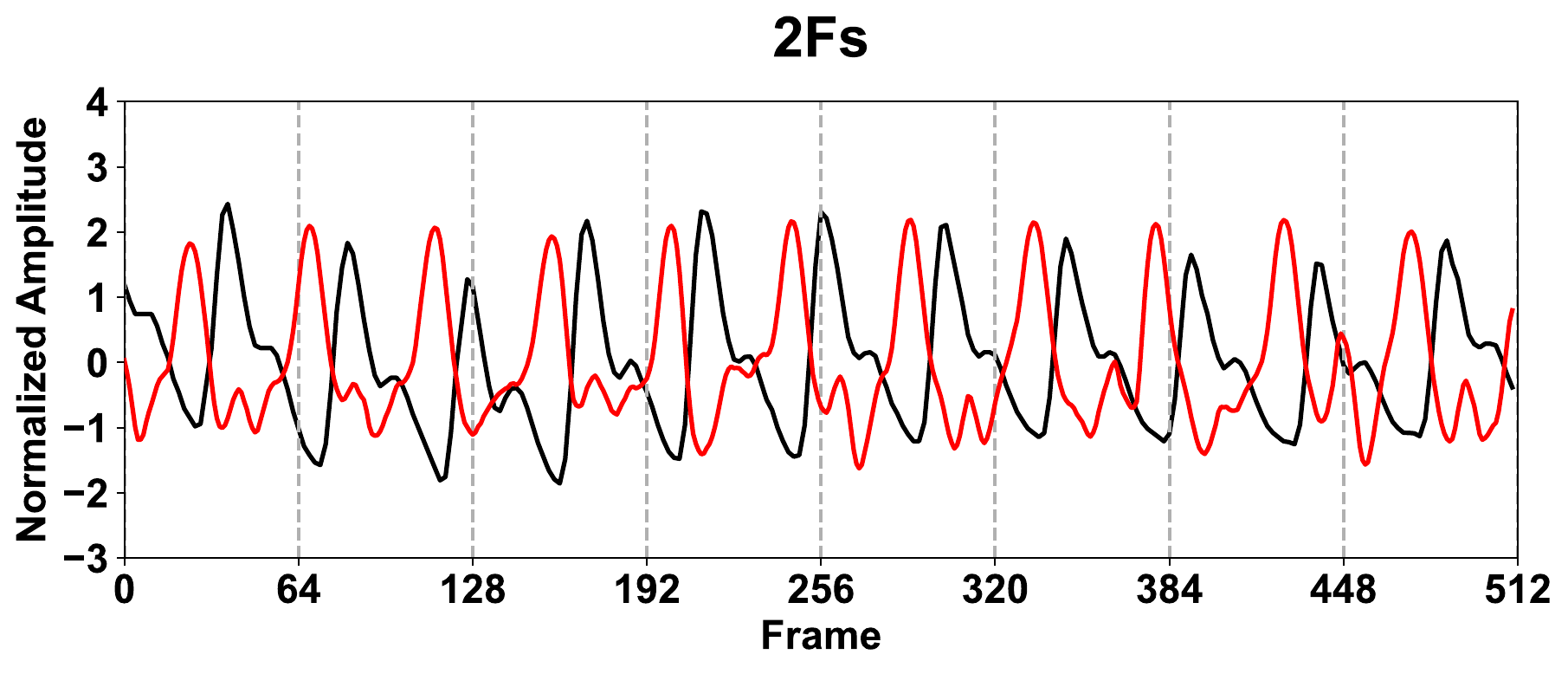}
    }\vspace{-0.18cm}

    \subfloat[Contrast-Phys \cite{Contrast-Phys}]{
        \includegraphics[width=0.325\textwidth]{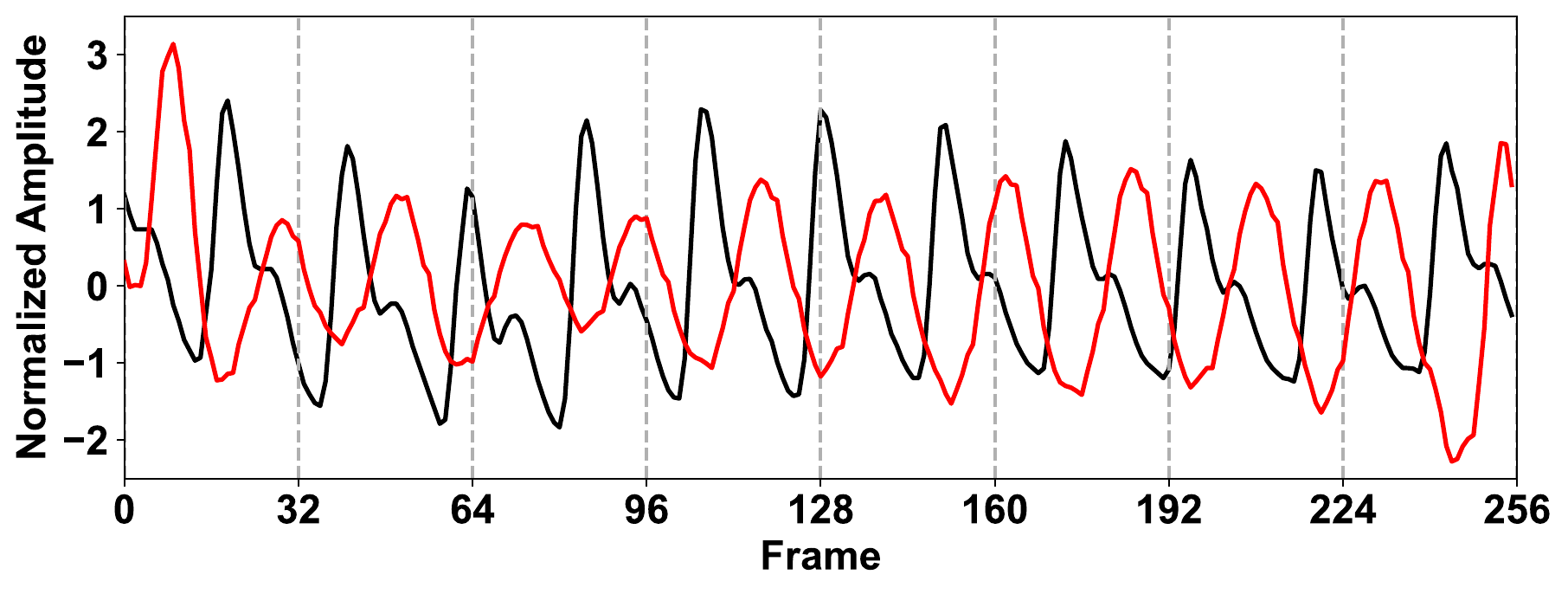}
        \includegraphics[width=0.325\textwidth]{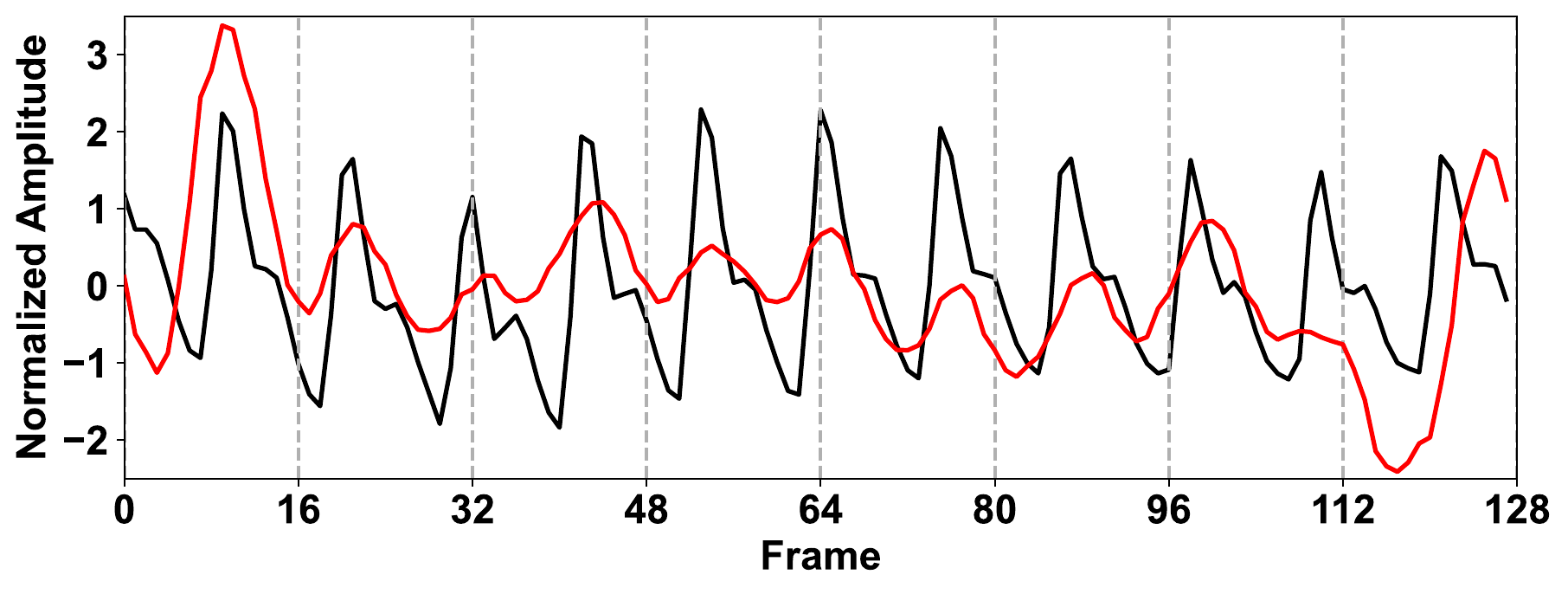}
        \includegraphics[width=0.325\textwidth]{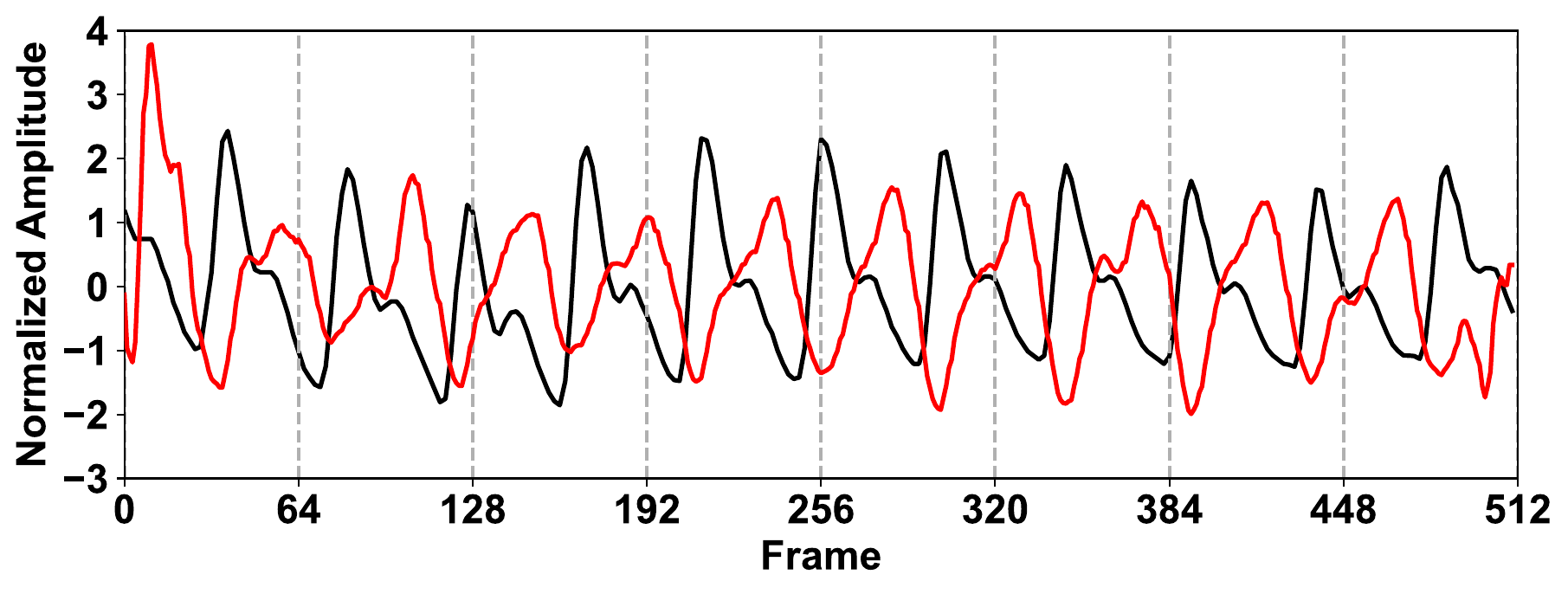}
    }\vspace{-0.18cm}

    \subfloat[RemotePPG \cite{RemotePPG}]{
        \includegraphics[width=0.325\textwidth]{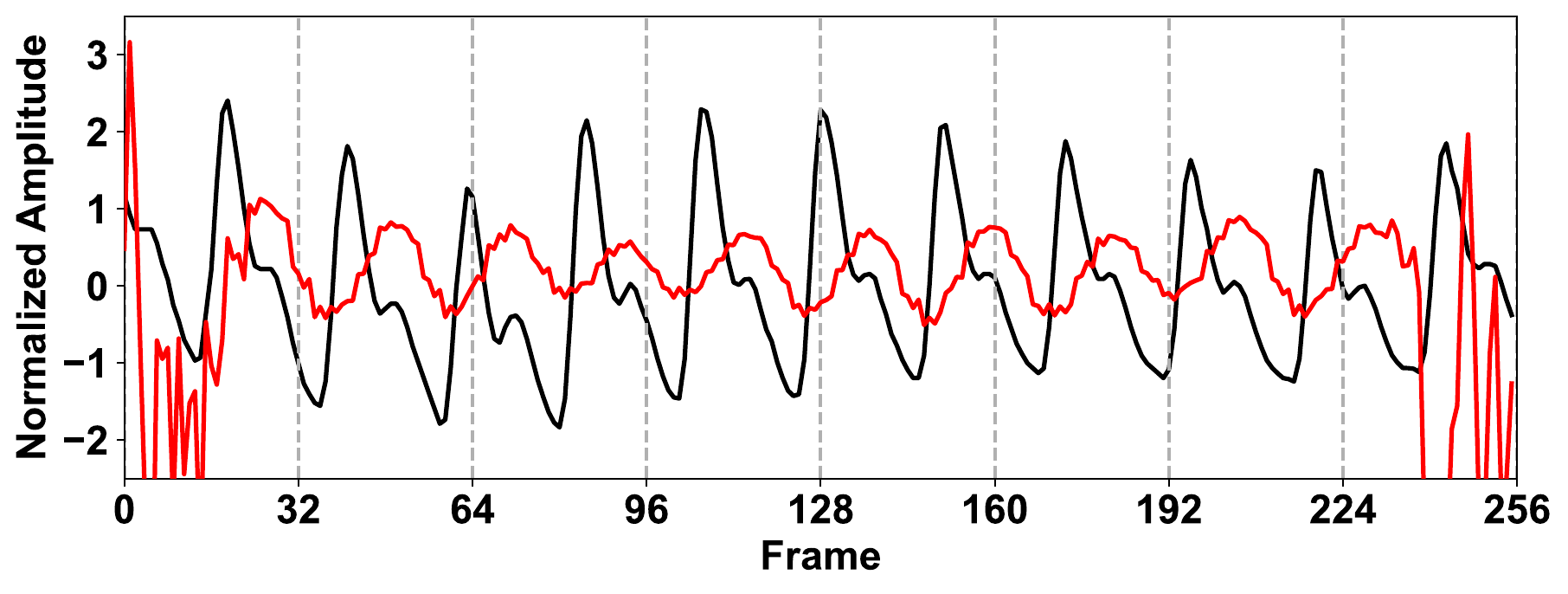}
        \includegraphics[width=0.325\textwidth]{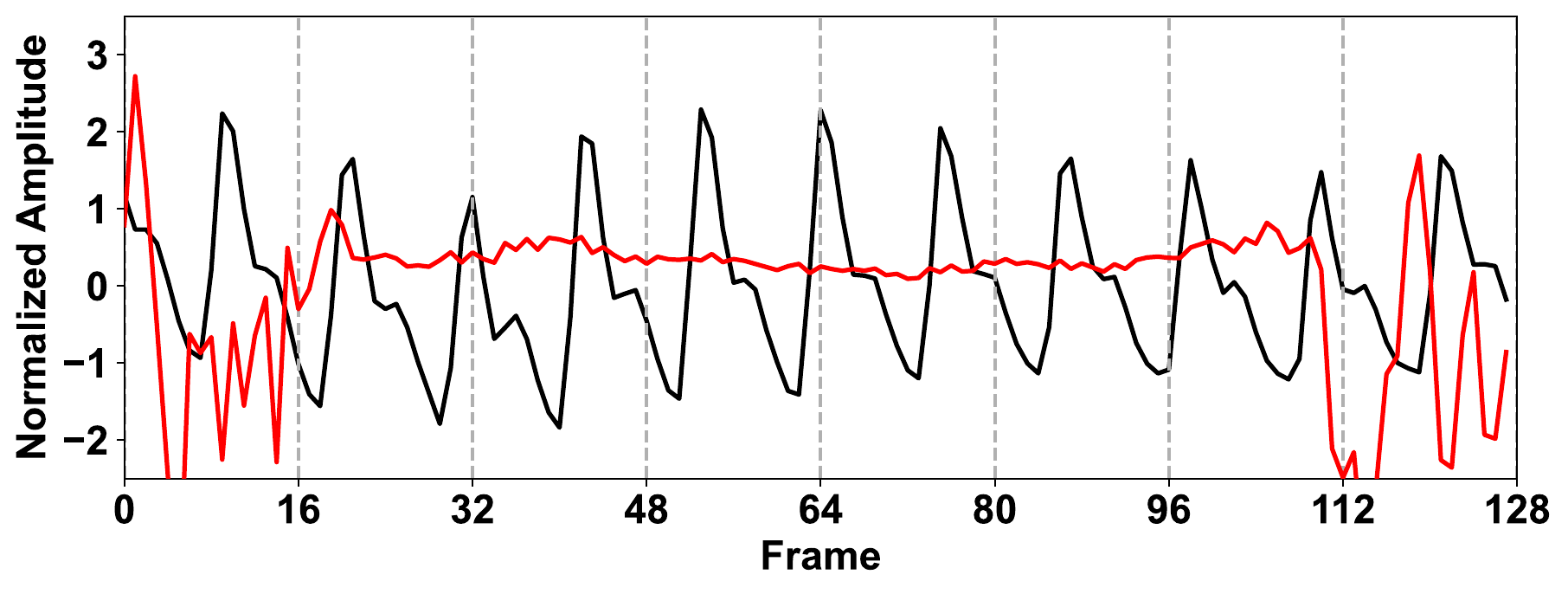}
        \includegraphics[width=0.325\textwidth]{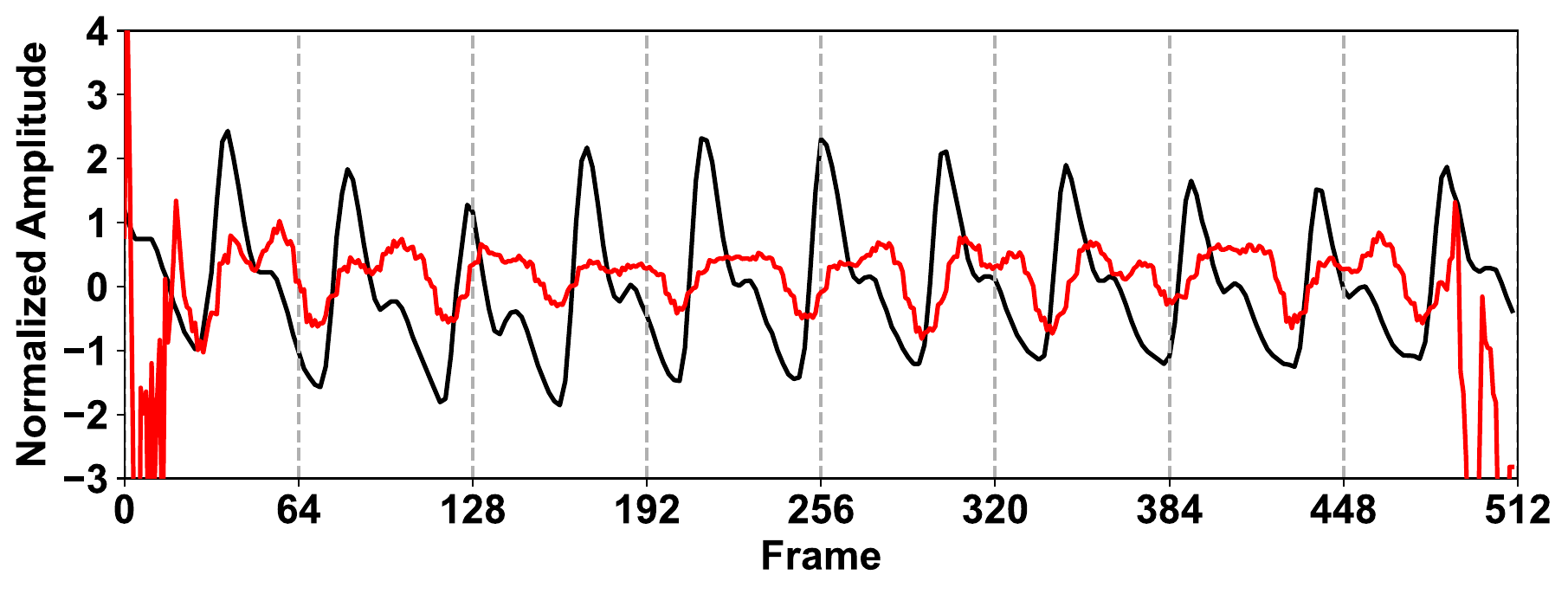}
    }\vspace{-0.18cm}

    \subfloat[ICA \cite{ICA}]{
        \includegraphics[width=0.325\textwidth]{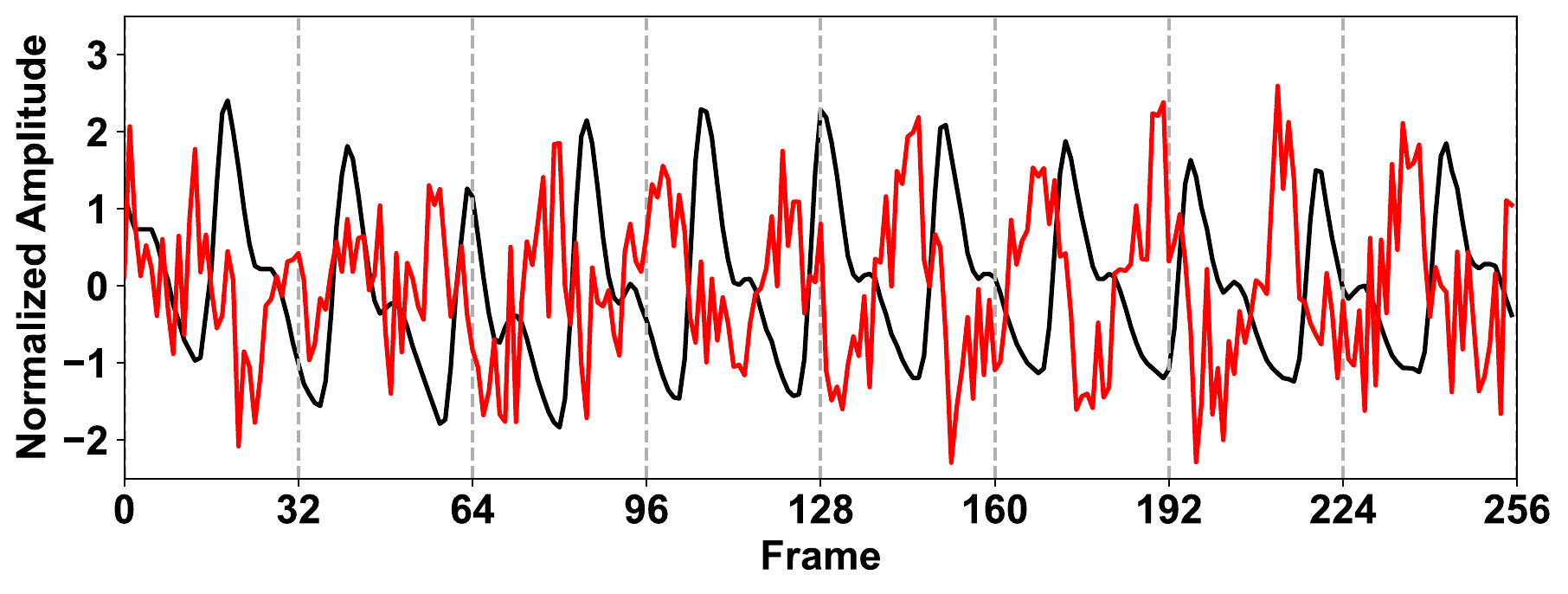}
        \includegraphics[width=0.325\textwidth]{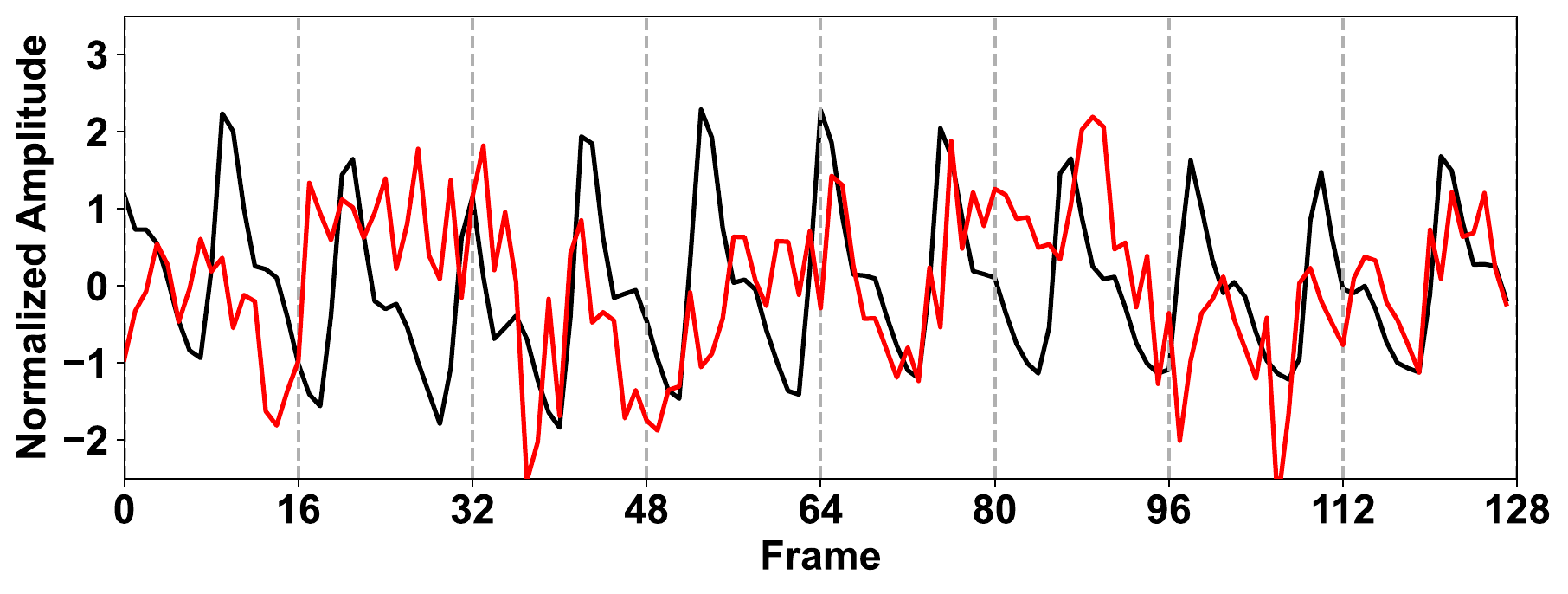}
        \includegraphics[width=0.325\textwidth]{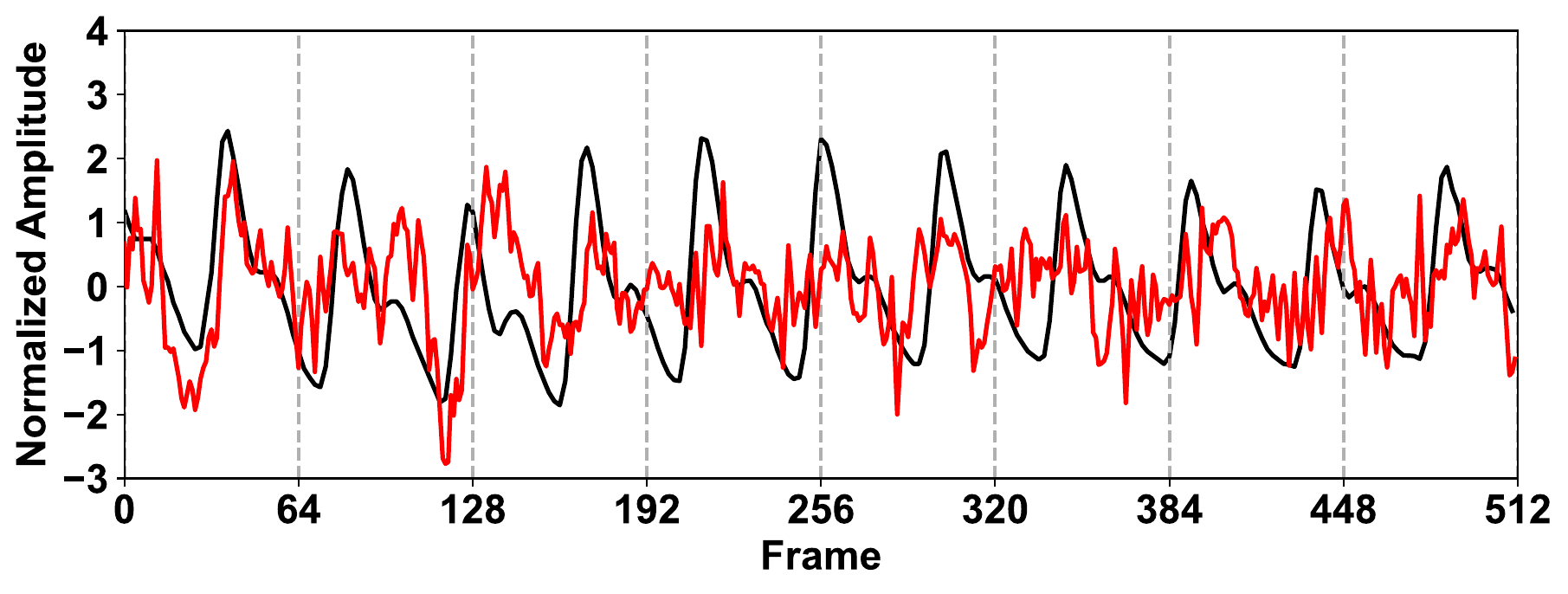}
    }
    \caption{The rPPG signal predictions on the first subject in the test set of the PURE dataset. We input the same clip with different sampling rates, namely $F_s$ (left column), $F_s/2$ (middle column), and $2F_s$ (right column), to the same model. The red lines represent the rPPG predictions, while the black lines represent the corresponding BVP labels. Note that all results are derived directly from the model's output without any filtering.} 
    \label{PURE visualization}
\end{figure*}

\subsection{Comparison with the State-of-the-art}
\subsubsection{Intra-dataset Testing}
Firstly, we perform intra-dataset evaluations for HR estimation on three small-scale benchmarks. As demonstrated in Table \ref{SSL result on small datasets}, our framework outperforms all the signal-based and learning-based unsupervised methods with a substantial improvement. Here, ``L3'' denotes that the $S^3M$ contains three TS blocks, i.e., the temporal similarity pyramid consists of three layers. In the PURE dataset, SSPD-L3 achieves higher performance than the current state-of-the-art supervised model, while in the UBFC-rPPG and MR-NIRP datasets, SSPD-L3 exhibits comparable results to the state-of-the-art supervised model. For a more intuitive illustration of the performance of our method, we first reproduce three competitive unsupervised baselines: Contrast-Phys \cite{Contrast-Phys}, RemotePPG \cite{RemotePPG}, and ICA \cite{ICA} (the best signal-based method), and generate regression and Bland-Altman plots on the UBFC-rPPG dataset.
Fig. \ref{UBFC visualization} shows that our framework displays the strongest correlation with the ground truth and the lowest standard deviation of errors, indicating that SSPD achieves the most precise and stable unsupervised HR measurement. 
Besides, in Fig. \ref{PURE visualization}, we showcase the rPPG estimation obtained from the first subject in the test set on the PURE dataset. It can be easily observed that the output of the proposed SSPD framework maintains a consistent rhythm with the BVP label across different sampling rates. Particularly, in a low sampling rate, SSPD exhibits robustness to information reductions, resulting in highly confident output peaks that are synchronized with the label and result in precise HR estimation. Meanwhile, in a high sampling rate, SSPD can estimate accurate rhythm while simultaneously capturing more subtle waveform dynamics, such as the diastolic peaks \cite{IAN}. This toy experiment demonstrates that our SSPD framework outperforms other unsupervised baselines, especially in the presence of sampling rate jitter. The efficacy of SSPD can be attributed to its capacity to acquire self-similar physiological features across multi-scale and long-distance intervals, which enables robustness to frequency perturbations and accurate prediction of extreme heart rates. Note that all waveforms are performed directly on the models' output without any filtering, and the observed phase shift between the predicted rPPG and the BVP label is attributed to the lack of manual alignment.

Next, we report the HR estimation results in the challenging VIPL-HR dataset in Table \ref{SSL VIPL result}. It should be highlighted that the previous unsupervised baseline \cite{ViViT} utilized a distinct test protocol, where they only assessed the performance on fold-1 of the VIPL-HR dataset, and each clip was limited to 160 frames to predict nearly instantaneous heart rate. 
To ensure a fair comparison, we follow the aforementioned protocol and present the performance of the SSPD-L3-160 model, whereas the SSPD-L3 model follows the strategy in line with the \cite{PhysFormer}. 
Noteworthy, ViViT \cite{ViViT} used the linear probing strategy to fine-tune their unsupervised model with label, while all of our results are obtained without any supervision. Furthermore, we reproduce the unsupervised state-of-the-art model \cite{Contrast-Phys, RemotePPG} on the VIPL-HR dataset based on open-source codes. As shown in Table \ref{SSL VIPL result}, our SSPD-L3 model significantly outperforms RemotePPG and Contrast-Phys using the same test strategy, indicating that the self-similarity prior facilitates the model to effectively capture physiological features, especially in large and complex datasets. 
In addition, we visualize the error distributions of 9 different tasks across all subjects in the VIPL-HR dataset.
The boxplot in Fig. \ref{VIPL plot} reveals that SSPD has notably lower errors compared to the unsupervised state-of-the-art model \cite{Contrast-Phys} and performs comparably to the supervised state-of-the-art model \cite{Dual-GAN} in most scenarios, especially in the task1: stable scenario, task4: bright scenario, task6: long distance scenario, and task8: phone stable scenario \cite{VIPL}. 

\begin{table}
    \centering
    \caption{The compute cost and running speed compared to the state-of-the-art end-to-end methods. The last column reports the HR estimation performance on the PURE dataset, and the best results are highlighted in \textbf{bold}. The $^\bigstar$ indicates the incorporation of the frame difference calculation process.}
    \label{Time cost}
    \resizebox{1.00                                                             \columnwidth}{!}{
    \setlength{\tabcolsep}{1.2mm}{}
    \begin{tabular}{clcccc}
    \toprule[1pt]
    \begin{tabular}[c]{@{}c@{}}Method\\ Types\end{tabular} & \multicolumn{1}{c}{Methods} & \multicolumn{1}{c}{Params} & GFLOPs & Time & \begin{tabular}[c]{@{}c@{}}MAE$\downarrow$ \end{tabular} \\ \midrule
    \multirow{3}{*}{Supervised}                            & PhysNet \cite{PhysNet}                     & \textbf{0.77M}             & 131.47     & \textbf{$\times$1.00}       & 2.10 \\
                                                           & CAN$^\bigstar$ \cite{CAN}                         & 7.50M                      & 225.01     & $\times$3.19                & 1.27 \\
                                                           & PhysFormer \cite{PhysFormer}                  & 7.38M                      & 94.89      & $\times$2.91                &  -   \\ \midrule
    \multirow{3}{*}{Unsupervised}                          & RemotePPG \cite{RemotePPG}                  & 0.86M                      & 256.68     & $\times$1.88                & 2.30 \\
                                                           & Contrast-Phys \cite{Contrast-Phys}               & 0.86M                      & 256.68     & $\times$1.88                & 1.00 \\
                                                           & \cellcolor[HTML]{FFE5E5}SSPD$^\bigstar$  & \cellcolor[HTML]{FFE5E5}0.78M  & \cellcolor[HTML]{FFE5E5}\textbf{84.79}   & \cellcolor[HTML]{FFE5E5}\textbf{$\times$1.00}      & \cellcolor[HTML]{FFE5E5}\textbf{0.53}  \\ \bottomrule[1pt]
    \end{tabular}
    }
    \vspace{-5pt}
\end{table}

\subsubsection{Cross-dataset Testing}
We conduct cross-dataset evaluations to exhibit the generalization of our SSPD framework. 
In Table \ref{SSL result on small datasets}, we present the HR estimation results obtained by selecting the best fold trained on VIPL-HR and testing it on PURE, UBFC-rPPG, and MR-NIRP. We directly compare them to the intra-dataset results, and SSPD significantly outperforms other unsupervised baselines in cross-dataset testing. 
The cross-dataset evaluations demonstrate that SSPD can learn self-similarity as a domain-invariant \cite{NEST, DOHA} characteristic inherent in various inputs. 
Consequently, SSPD is insensitive to distributional differences among datasets compared to contrastive learning baselines.

\subsubsection{Inference Cost}
To demonstrate the effectiveness of the proposed self-similarity-aware network, particularly the design of the $S^3M$, we first reproduce the end-to-end rPPG baselines based on open-source codes \cite{PhysFormer, Contrast-Phys, RemotePPG, toolbox},
then we calculate the computational complexity and actual inference time under the same hardware environment, as shown in Table \ref{Time cost}. The input size for all tests is 300$\times$128$\times$128 and the running time is obtained by averaging 10 independent tests. It should be noted that the running time additionally includes the frame difference calculation in both CAN \cite{CAN} and SSPD for a fair comparison. 
According to Table \ref{Time cost}, our framework has the lowest compute cost and the fastest running speed during inference while still achieving the best performance. Specifically, compared to the current unsupervised state-of-the-art model \cite{Contrast-Phys}, our method has almost 3$\times$ fewer GFLOPs and 2$\times$ less inference time, while the end-to-end supervised state-of-the-art model \cite{PhysFormer} has nearly 10$\times$ parameters and 3$\times$ more inference time than ours. Thus, our method is carbon-neutral, making it suitable for embedded and edge computing platforms.

\subsection{Ablation Studies}\label{ablation section}
In this section, we conduct comprehensive ablation studies on the UBFC-rPPG dataset to investigate the effectiveness of the SSPD framework.

\begin{table}
    \centering
    \caption{Ablation studies on hierarchical self-distillation and periodicity regularizations in our SSPD framework. The best results are highlighted in \textbf{bold}.}
    \label{ablation on loss}
    \setlength{\tabcolsep}{1.2mm}{}
    \resizebox{1.\columnwidth}{!}{
    \begin{tabular}{c|cccc|ccc}
    \toprule[1pt]
    \multirow{2}{*}{Methods} & \multirow{2}{*}{$\mathcal{L}_{SD}$} & \multicolumn{2}{c}{$\mathcal{L}_{Distill}$} & \multirow{2}{*}{$\mathcal{L}_{SNR}$} & \multirow{2}{*}{\begin{tabular}[c]{@{}c@{}}MAE$\downarrow$ \end{tabular}} & \multirow{2}{*}{\begin{tabular}[c]{@{}c@{}}RMSE$\downarrow$\end{tabular}} & \multirow{2}{*}{R$\uparrow$} \\ \cline{3-4}
                             &                    & $\mathcal{L}_{RPD}$         & $\mathcal{L}_{TSPD}$        &                     &                                                                      &                                                                       &                    \\ \midrule
                             w/o TSPD             & -  & \checkmark  & -           & -     & 3.50        & 10.25       & 0.803    \\
                             w/o RPD              & -  & -           & \checkmark  & -     & 0.85        & 1.24        & 0.995    \\ 
                             $\mathcal{L}_{Distill}$ only    & -           & \checkmark  & \checkmark  & -             & 0.76       & 1.21        & 0.995    \\ \midrule 
                             w/o $\mathcal{L}_{SD}$          & -           & \checkmark  & \checkmark  & \checkmark    & 0.66       & 0.91        & 0.997    \\
                             w/o $\mathcal{L}_{SNR}$         & \checkmark  & \checkmark  & \checkmark  & -             & 0.78       & 1.14        & 0.995    \\
                             default              & \checkmark  & \checkmark  & \checkmark  & \checkmark    & \textbf{0.50}       & \textbf{0.71}        & \textbf{0.998}    \\ \bottomrule[1pt]
    \end{tabular}
  }
\end{table}

\begin{figure}
    \centering
    \subfloat[momentum rate $\rho$]{\includegraphics[width=0.23\textwidth]{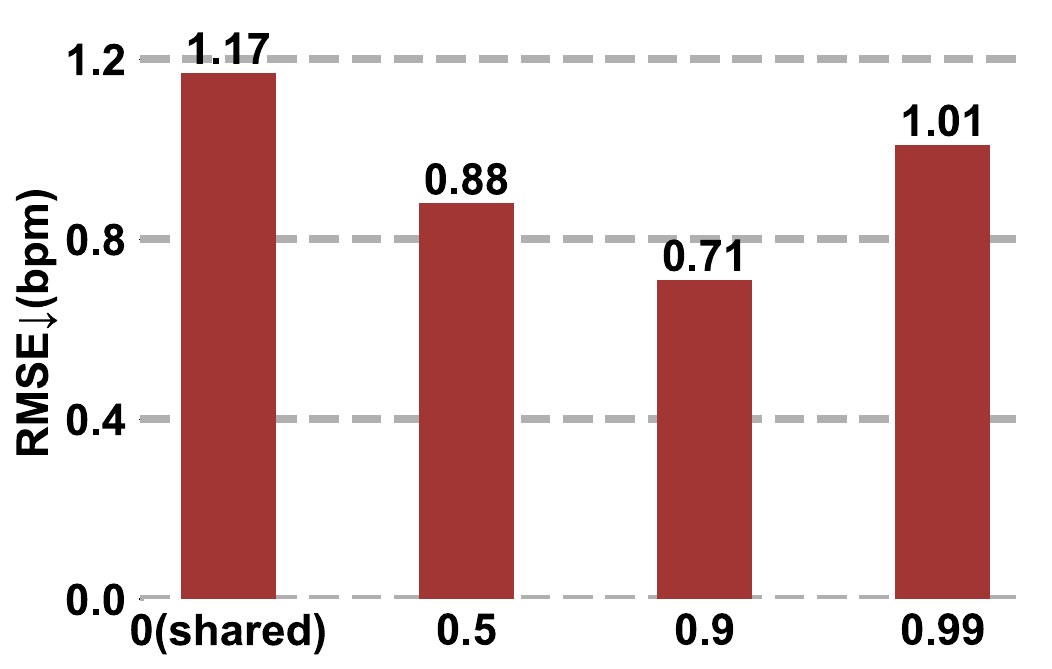}}
    \subfloat[mask ratio $p$]{\includegraphics[width=0.25\textwidth]{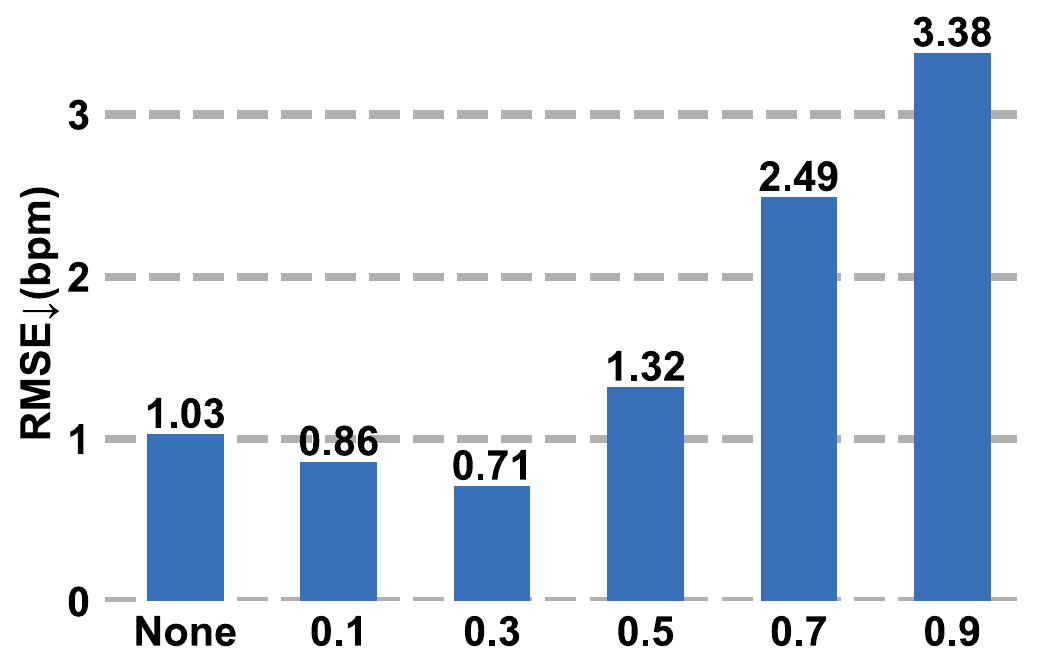}}
    \caption{The effects of hyperparameters, including (a) momentum rate $\rho$ and (b) mask ratio $p$, on the proposed SSPD framework.} 
    \label{Ablations on mask ratio and momentum rate}
\end{figure}

\begin{figure}
    \centering
    \includegraphics[width=0.48\textwidth]{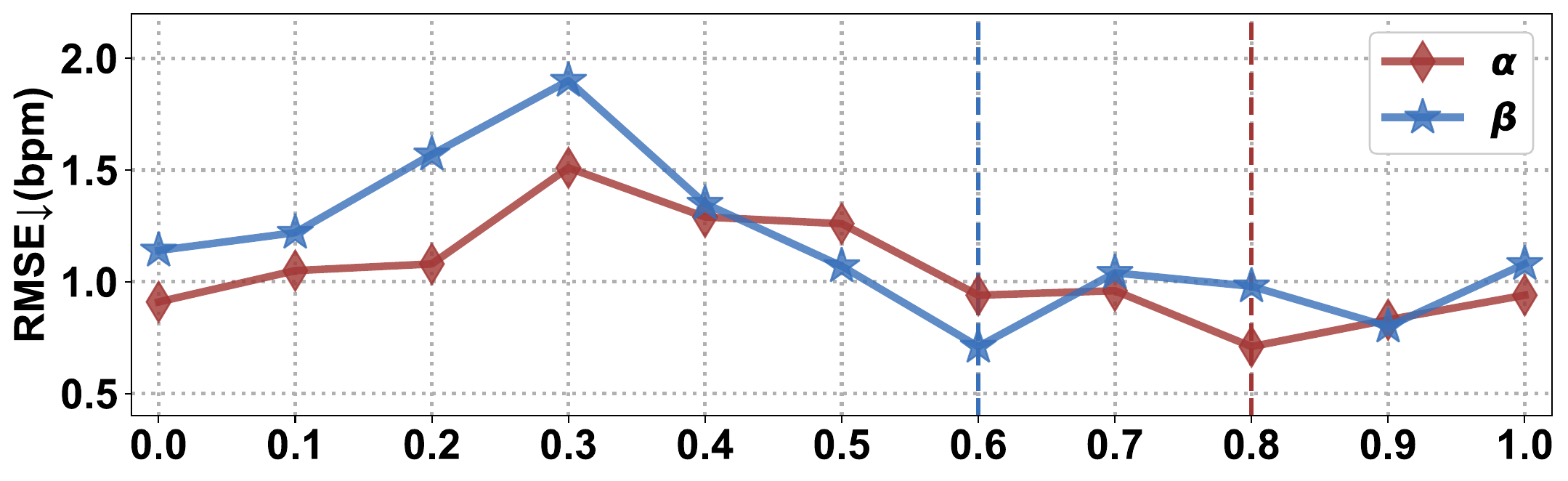}
    \caption{Impacts of the hyperparameters (a) $\alpha$, (b) $\beta$ on the tradeoff between two periodicity regularizations.} 
    \label{Ablations on alpha and beta}
\end{figure}

\subsubsection{Hierarchical Self-distillation}
First, we investigate the significance of the two distilled knowledge in our SSPD framework, corresponding to the self-similarity-aware learning and rPPG signal decoupling processes.
The upper part of Table \ref{ablation on loss} shows that dropping the Temporal Similarity Pyramid Distillation (TSPD) results in significant performance degradation, as it disables the entire $S^3M$, 
making it infeasible for SSPD to learn the self-similarity knowledge via SSM/SSW distillation. Interestingly, reasonable heart rate estimations are still achieved without RPPG Prediction Distillation (RPD). This phenomenon can be attributed to the shared self-similarity between the temporal similarity pyramid and the predicted rPPG signal, indicating that the backbone has captured reliable physiological representations, which have the same temporal rhythms as the rPPG signal.
Further visualization of the self-distillation dynamic in SSPD will be provided in Sec. \ref{Self-distillation Dynamic}.
Next, as demonstrated in the lower part of Table \ref{ablation on loss}, 
incorporating both SD and SNR regularization results in improved performance, emphasizing the importance of exploiting periodicity in SSM and SSW. 
We further perform a grid search for hyperparameters $\alpha$ and $\beta$, as presented in Eqn. \ref{total loss}. 
Based on the results in Fig. \ref{Ablations on alpha and beta}, we set $\alpha$ and $\beta$ to 0.8 and 0.6 by default, respectively.

Besides, Fig. \ref{Ablations on mask ratio and momentum rate}(a) illustrates the importance of the momentum rate $\rho$ in EMA, as defined in Eqn. \ref{EMA}. Results show that a relatively small momentum rate benefits model learning. Empirically, an initial momentum rate of 0.9 is found to be optimal for small-scale datasets, while 0.99 is more suitable for the larger VIPL-HR dataset. 
Noteworthy is that SSPD is very flexible in target model design. 
SSPD remains effective even when the weights of the online and target models are shared directly, 
following the previous unsupervised rPPG baselines \cite{Contrast-Phys, RemotePPG, ViViT}. 

\begin{table}
    \centering
    \caption{Ablation studies on the physical-prior embedded augmentation. The best results are highlighted in \textbf{bold}.}
    \label{ablation on views}
    \setlength{\tabcolsep}{1.5mm}{}
    \resizebox{1.\columnwidth}{!}{
      \begin{tabular}{c|ccc}
        \toprule
        Methods   & \tabincell{c}{MAE$\downarrow$} & \tabincell{c}{RMSE$\downarrow$} & \tabincell{c}{R$\uparrow$}  \\
        \midrule
        w/o LGA    & 1.09       & 2.15        & 0.981    \\
        w/o frame difference mapping in MDM   & 4.15    & 6.10     & 0.842    \\
        w/o random masking in MDM   & 0.63    & 1.03     & 0.995    \\
        default   & \textbf{0.50}       & \textbf{0.71}        & \textbf{0.998}   \\
        \bottomrule
      \end{tabular}
    }
    \vspace{-10pt}
\end{table}

\begin{table}
    \vspace{-0.4cm}
    \centering
    \caption{Ablation studies on the architecture of the TS block. The best results are highlighted in \textbf{bold}.}
    \label{ablation on TS block}
    \setlength{\tabcolsep}{1.2mm}{}
    \resizebox{1.\columnwidth}{!}{
      \begin{tabular}{c|ccc}
        \toprule
        Methods   & \tabincell{c}{MAE$\downarrow$} & \tabincell{c}{RMSE$\downarrow$} & \tabincell{c}{R$\uparrow$}  \\
        \midrule
        w/o residual connection                             & 1.07  & 2.13    & 0.980    \\
        w/o dot-product attention                           & 0.81  & 1.21    & 0.994    \\
        w/o projection                                      & 0.55  & 0.78    & 0.997    \\ \midrule
        dot-product attention $\rightarrow$ self-attention  & 0.71  & 1.03    & 0.995    \\
        multi-head $\rightarrow$ single-head    & 0.69   & 1.00   & 0.996    \\
        default                                             & \textbf{0.50}       & \textbf{0.71}        & \textbf{0.998}   \\
        \bottomrule
      \end{tabular}
    }
\end{table}

\begin{table}
    \vspace{-0.6cm}
    \centering
    \caption{Ablation studies on the input shape of the SSPD framework, the number of layers, and the token size in $S^3M$. The best results are highlighted in \textbf{bold}.}
    \label{ablation on pyramid}
    \setlength{\tabcolsep}{1.7mm}{}
    \resizebox{1.\columnwidth}{!}{
      \begin{tabular}{c|c|c|ccc}
        \toprule[1pt] 
        Methods     & Input Size  & Token Size  & \tabincell{c}{MAE$\downarrow$} & \tabincell{c}{RMSE$\downarrow$} & \tabincell{c}{R$\uparrow$}  \\
        \midrule
        SSPD-L1      & 300$\times$128$\times$128   & \makecell{$\left[292\times256\right]$}                                                                                                 & 0.73       & 1.46        & 0.991    \\ \midrule
        SSPD-L2      & 300$\times$128$\times$128   & \makecell{$\left[292\times256\right]$ \\ $\left[286\times256\right]$}                                                                  & 0.60       & 0.94        & 0.996    \\ \midrule
        SSPD-L3      & 300$\times$128$\times$128   & \makecell{$\left[292\times128\right]$ \\ $\left[286\times128\right]$ \\ $\left[282\times128\right]$}                                   & 0.55       & 0.79        & \textbf{0.998}    \\ \midrule
        SSPD-L3      & 300$\times$128$\times$128   & \makecell{$\left[292\times256\right]$ \\ $\left[286\times256\right]$ \\ $\left[282\times256\right]$}                                   & \textbf{0.50}       & \textbf{0.71}        & \textbf{0.998}    \\ \midrule
        SSPD-L3      & 300$\times$128$\times$128   & \makecell{$\left[292\times512\right]$ \\ $\left[286\times512\right]$ \\ $\left[282\times512\right]$}                                   & \textbf{0.50}       & 0.72       & \textbf{0.998}    \\ \midrule
        SSPD-L3      & 300$\times$128$\times$128   & \makecell{$\left[296\times256\right]$ \\ $\left[144\times256\right]$ \\ $\left[68\times256\right]$}                                    & 0.69       & 1.01        & 0.995    \\ \midrule
        SSPD-L3      & 160$\times$128$\times$128   & \makecell{$\left[152\times256\right]$ \\ $\left[146\times256\right]$ \\ $\left[142\times256\right]$}                                   & 0.54       & 0.79        & 0.997    \\ \midrule
        SSPD-L3      & 300$\times$96$\times$96     & \makecell{$\left[292\times256\right]$ \\ $\left[286\times256\right]$ \\ $\left[282\times256\right]$}                                   & 0.60       & 0.85        & 0.996    \\ \midrule
        SSPD-L4      & 300$\times$128$\times$128   & \makecell{$\left[292\times256\right]$ \\ $\left[286\times256\right]$ \\ $\left[282\times256\right]$ \\ $\left[280\times256\right]$}    & 0.54       & 0.76        & 0.997    \\
        \bottomrule[1pt] 
      \end{tabular}
    }
    \vspace{-8pt}
\end{table}

\subsubsection{Physical-prior Embedded Augmentation}
The results presented in Table \ref{ablation on views} exhibit the impact of two physical-prior embedded augmentation strategies. Specifically, Local-Global Augmentation (LGA) is crucial in enhancing the local-global response, thus improving modeling capability at different spatial resolutions. On the other hand, Masked Difference Modeling (MDM) further increases the deviation between two augmented views and benefits self-similarity-aware learning. 
It is worth noting that frame difference mapping is a crucial component of MDM (-3.65 bpm MAE), with its ability to enable the model to concentrate on motion information. Nevertheless, incorporating random masking in MDM can further improve the estimation accuracy based on the aforementioned techniques.
In addition, we search for the optimal mask ratio $p$, as shown in Fig \ref{Ablations on mask ratio and momentum rate}(b).
The results indicate that a mask ratio of 0.3 yields the best performance, while overly high mask ratios result in dramatic degradation.

\subsubsection{TS Block}
In Table \ref{ablation on TS block}, we conduct a comprehensive analysis on the components of the TS block. The first two rows show that the residual connection and dot-product attention are the most crucial elements of the block. That is because residual connection enhances the capturing of self-similar representations across multiple time scales, while the attention mechanism enhances the long-distance physiological features learning within each specific time scale. Furthermore, dot-product attention, as defined in Eqn. \ref{DotProduct Attention}, 
and multi-head attention outperform self-attention and single-head attention, respectively. Finally, the projection module following the multi-head attention further boosts the performance by fusing the information from different heads.

\subsubsection{Self-similarity-aware Network}
In our ablation on the implementation of the self-similarity-aware network, we explore multiple factors, including the input shape of the SSPD framework, as well as the number of layers and token size in the $S^3M$. The results are illustrated in Table \ref{ablation on pyramid}, 
which can be summarized as follows: (a) The incorporation of multiple time scales improves the performance of the network. By comparing the results obtained with different numbers of layers in the temporal similarity pyramid, we observe that the performance boosts with an increasing number of layers, and then reaches saturation at SSPD-L4. Moreover, the computational cost of training significantly increases with the number of blocks, hence we use SSPD-L3 by default. (b) The self-similarity-aware network is insensitive to the dimension of token embeddings. As evidenced by the results in rows three to five of Table \ref{ablation on pyramid},
there is little variation in the MAE, RMSE, and R metrics despite an increase in dimension from 128 to 512. (c) A dramatic change in time scales leads to degradation. To address this issue, we employ an adaptive average pooling layer in the tokenizer of each TS block, where the time scale is reduced by half at each layer of the pyramid. As shown in the sixth row of Table \ref{ablation on pyramid},
the MAE increases from 0.5 to 0.69 due to the dramatic changes in time scales. The reduced granularity of self-similar representations in the time domain can result in decreased accuracy in rPPG estimation. (d) The self-similarity-aware network exhibits robustness to variations in input shape. Specifically, we evaluate the impact of reducing the temporal length of the input clips from 300 to 160 and the spatial size from 128$\times$128 to 96$\times$96. Our results show a slight decrease in performance due to the reduced shape in both space and time domains. Nonetheless, the SSPD framework maintains its state-of-the-art performance in unsupervised methods.

\begin{table}
    \centering
    \caption{Ablation studies on self-similarity-aware learning and rPPG signal decoupling strategies. The best results are highlighted in \textbf{Bold}}
    \label{policy}
    \setlength{\tabcolsep}{1mm}{}
    \resizebox{1.\columnwidth}{!}{
      \begin{tabular}{c|c|ccc}
        \toprule
        Process  & Methods   & \tabincell{c}{MAE$\downarrow$ \\} & \tabincell{c}{RMSE$\downarrow$ \\} & \tabincell{c}{R$\uparrow$}  \\
        \midrule
        \multirow{3}{*}{\begin{tabular}[c]{@{}c@{}}self-similarity-aware \\learning\end{tabular}}
                                    & w/o SSM and SSW ($\mathcal{L}_{MSE}$)  & 2.60   & 8.26   & 0.841    \\
                                    & w/ SSM, w/o SSW  & 0.97   & 1.68   & 0.988    \\
                                    & w/ SSM and SSW (default)   & \textbf{0.50}   & \textbf{0.71}   & \textbf{0.998}  \\ \midrule
        \multirow{2}{*}{\begin{tabular}[c]{@{}c@{}}rPPG signal \\ decoupling\end{tabular}}
                                    & predictor w/o stop-gradient  & 0.80   & 1.30    & 0.995    \\
                                    & predictor w/ stop-gradient (default)   & \textbf{0.50}    & \textbf{0.71}    & \textbf{0.998} \\
        \bottomrule
      \end{tabular}
    }
\end{table}

\subsubsection{Self-similarity-aware Learning and rPPG Signal Decoupling Strategies}
Based on Table \ref{ablation on loss}, we further present multiple self-similarity-aware learning and rPPG signal decoupling strategies in Table \ref{policy} to demonstrate the efficacy of SSM/SSW distillation and rPPG decoupling, as discussed in Sec \ref{Self-similarity-aware Network}.
First, we replace $\mathcal{L}_{TSPD}$ with $\mathcal{L}_{MSE}$, which directly distills the output tokens of the TS blocks (detailed in Fig. \ref{TS block}) rather than transforming them into temporal similarity pyramid. 
The results in the first row of Table \ref{policy} indicate a significant performance degradation, highlighting the critical role of self-similarity knowledge. 
Then, we disable SSW distillation from $\mathcal{L}_{TSPD}$ as well as $\mathcal{L}_{SNR}$ to investigate the efficacy of SSM knowledge. Compared to directly utilizing $\mathcal{L}_{MSE}$, SSM significantly enhances performance by incorporating the self-similarity prior. 
Building upon this, further exploration of periodicity in SSM yields the best performance.
Moreover, we conduct ablations on two rPPG signal decoupling strategies: updating the predictor along with the backbone or solely updating the predictor. 
Experimental results show that adding stop-gradient leads to a slightly better performance since simultaneously processing self-similarity-aware learning and rPPG signal decoupling in the backbone intensifies the learning burden in the early training stage, as discussed in Sec. \ref{Self-similarity-aware Network}.

\section{Discussion}\label{Discussion}
In this section, we will further discuss the key design in our SSPD framework, including the feasibility of the self-distillation paradigm and why the self-similarity prior achieves better performance than previous unsupervised work.

\subsection{Self-distillation Dynamic}\label{Self-distillation Dynamic}
\begin{figure}
    \centering
    \includegraphics[width=0.485\textwidth]{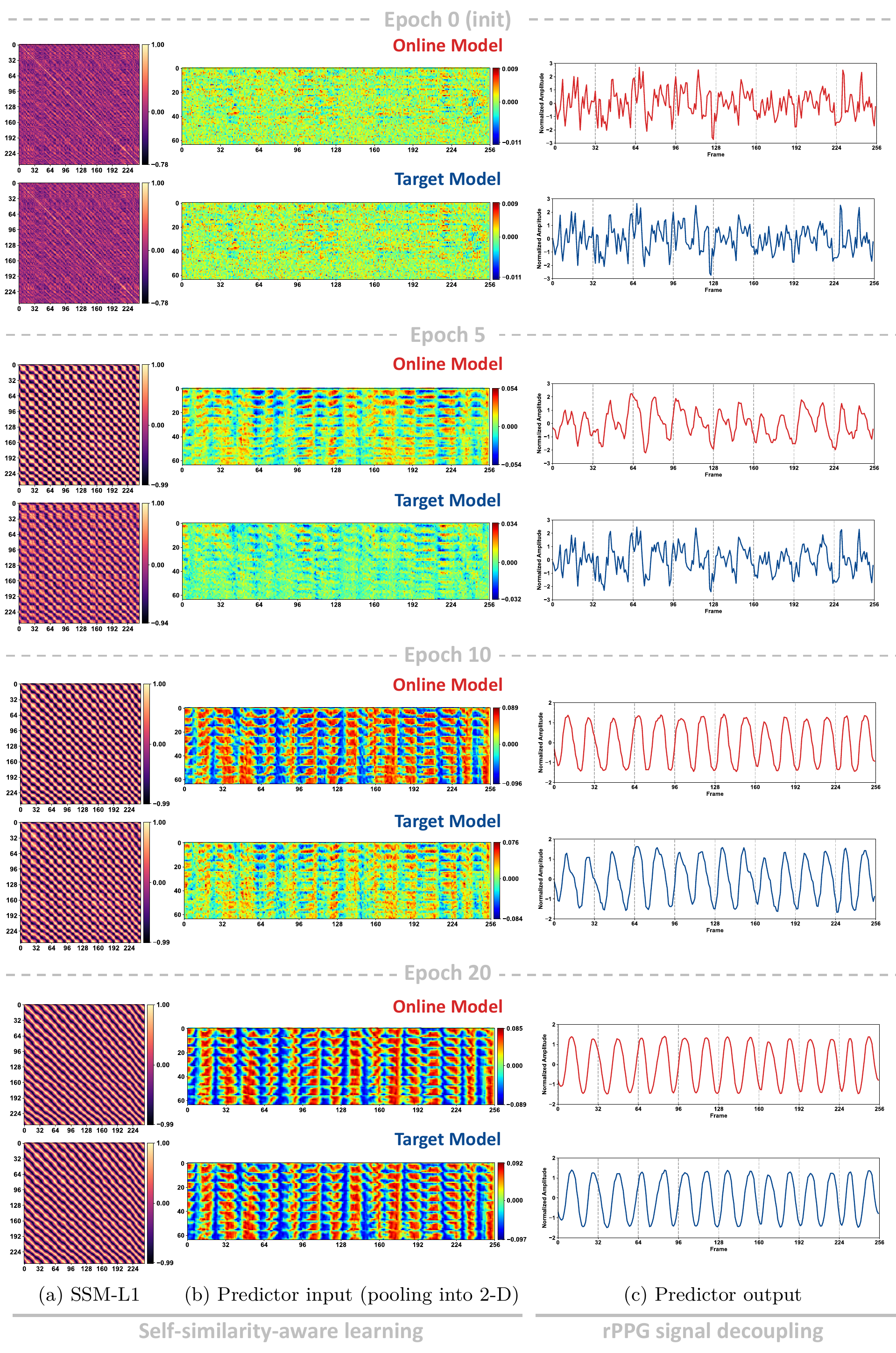}
    \caption{Self-distillation dynamic in SSPD framework. We visualize the SSM output from $S^3M$ (left), feature map output from the backbone (middle), and rPPG output from the predictor (right) at different trining epochs.
    It is evident that both the backbone and $S^3M$ can learn periodic physiological features through self-similarity-aware learning. Subsequently, based on the shared temporal self-similarity between facial video and physiological signal, the reliable decoupling of the rPPG signal is accomplished through the predictor module. Overall, the entire process is unsupervised.} 
    \label{dynamic}
\end{figure}

We visualize the training dynamic for self-distillation in SSPD in Fig. \ref{dynamic}. 
According to Algorithm \ref{Pseudocode}, both the online and target models start with the same initialization parameters and distill shared self-similarity knowledge from different input views throughout training. 
As an iterative ensemble of the online model, the target model gradually provides high-quality supervision while approaching the online model's parameters. 
In the self-similarity-aware learning process, depicted in Fig. \ref{dynamic}(a), 
$S^3M$ progressively learns the self-similarity relationship among tokens through $\mathcal{L}_{TSPD}$ and periodicity regularizations, propagating the self-similarity knowledge to the shallow backbone. 
In Fig. \ref{dynamic}(b), we visualize the averaged pooled output feature map of the backbone (i.e., the predictor input) as $\mathbb{R}^{256 \times T \times 8 \times 8} \rightarrow \mathbb{R}^{64 \times T}$ (detailed in Fig. \ref{backbone predictor module}), 
where the width of the resized map represents the temporal frame, and the height corresponds to the spatial receptive fields. 
It demonstrates that the backbone effectively captures the temporal self-similarity in facial videos, 
validating the efficacy of self-similarity-aware learning. 
Building upon this, as shown in Fig. \ref{dynamic}(c), 
we ultimately generate the rPPG waveform using $\mathcal{L}_{RPD}$ and $\mathcal{L}_{SNR}$ based on the output of the backbone.

\begin{figure}
    \centering
    \includegraphics[width=0.4\textwidth]{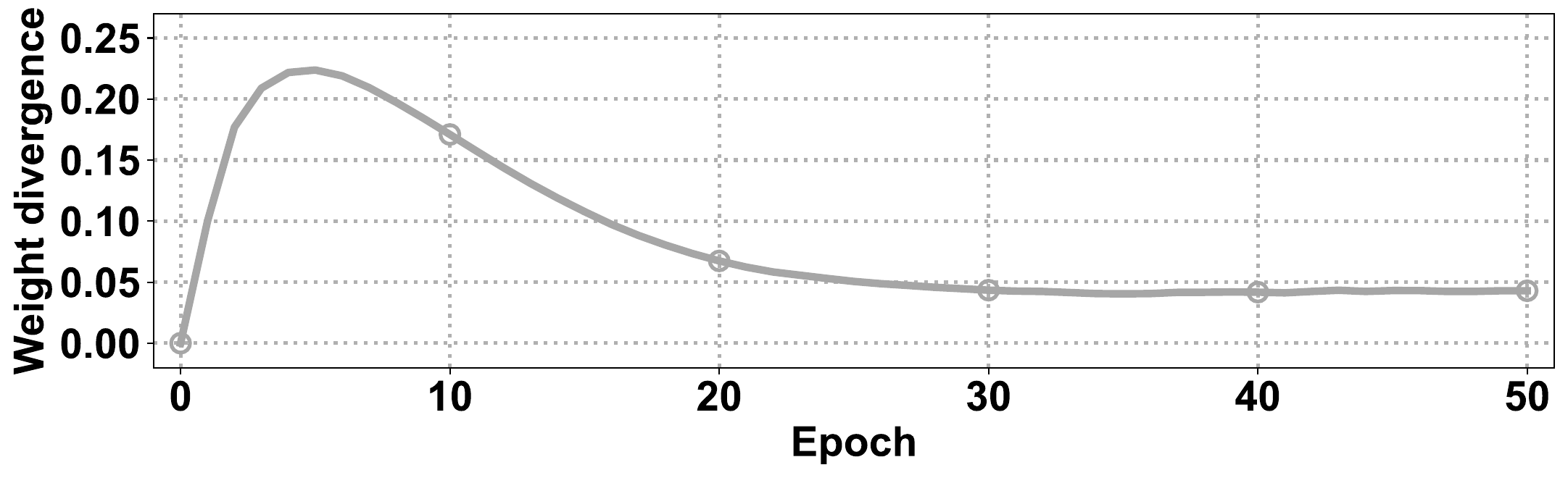}
    \caption{Weight divergence between online and target models at different training epochs on UBFC-rPPG.} 
    \label{weight div}
\end{figure}

\begin{table}
    \centering
    \caption{Performance comparison of online and target models on different datasets, where a lower MAE indicates a higher performance.}
    \label{SD}
    \resizebox{1.00\columnwidth}{!}{
    \setlength{\tabcolsep}{2.3mm}{}
    \begin{tabular}{c|cccc}
    \toprule
    Model Types & \tabincell{c}{UBFC-rPPG} & \tabincell{c}{PURE} & \tabincell{c}{MR-NIRP} & \tabincell{c}{VIPL-HR} \\ \midrule
    online model $\theta_\mathcal{O}$  & 0.53 & 0.52 & 2.74 & 6.09 \\
    target model $\theta_\mathcal{T}$  & 0.50 & 0.53 & 2.65 & 6.04 \\
    \bottomrule
    \end{tabular}
    }
\end{table}

Furthermore, we evaluate the weight divergence between the online and target models during training and the final performance of both to investigate the EMA-based parameter updating process \cite{DINO}.
First, we measure the weight divergence by calculating the average L2 distance of parameters (i.e., $\frac{1}{|\theta_{\mathcal{O}}|}\sum{\left\lVert \theta_{\mathcal{O}} - \theta_{\mathcal{T}} \right\rVert}_2$).
As shown in Fig. \ref{weight div}, online and target models start with the same parameters. 
Ultimately, driven by the EMA mechanism, the target model gradually converges towards the parameters of the online model. 
Then, we evaluate the performance of online and target models in Table \ref{SD}. Notably, both models demonstrate effectiveness during inference. 
Interestingly, the target model slightly outperforms the online model in most scenarios, potentially due to it is the ensemble of past iterations of the online model, which is more robust. 
Consequently, in all our experiments, we default to reporting the performance of the target model unless otherwise specified.

\subsection{Positive-pair similarity \& Self-similarity}

\begin{figure}[H]
    \centering
    \subfloat[Positive-pair similarity \cite{Contrast-Phys, RemotePPG}]{\includegraphics[width=0.48\textwidth]{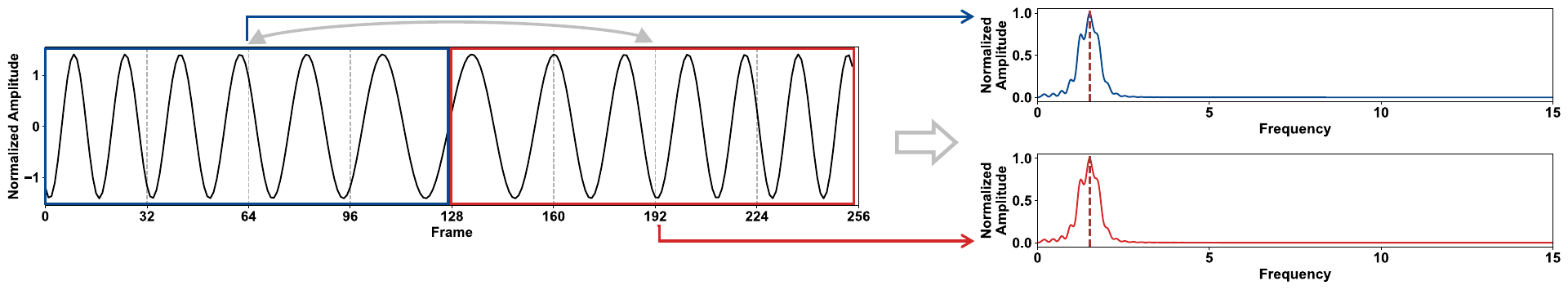}} \\ \vspace{-0.4cm}
    \subfloat[Self-similarity (ours)]{\includegraphics[width=0.48\textwidth]{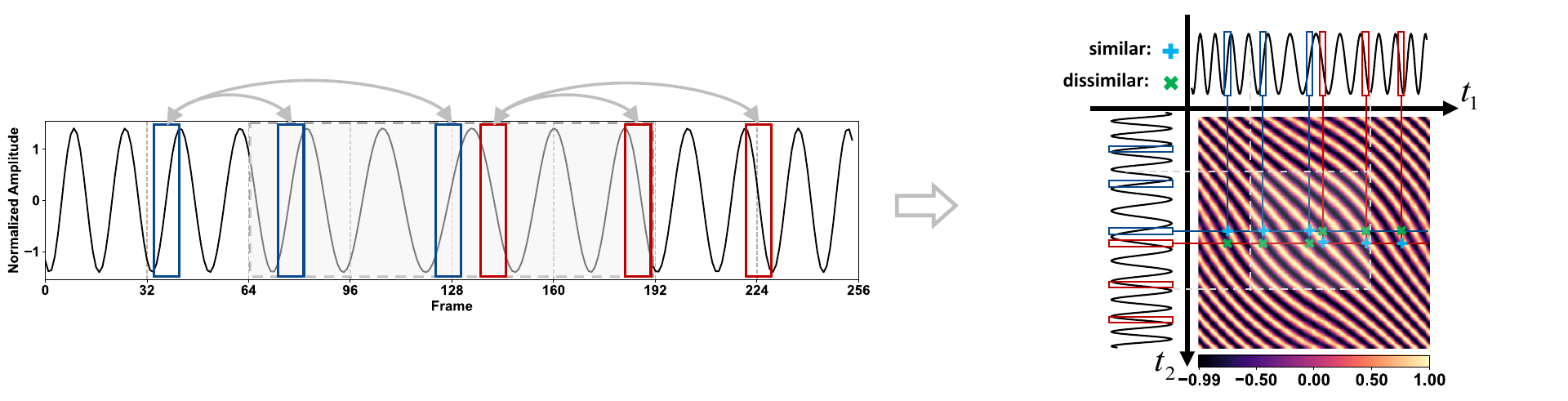}}
    \caption{The self-similarity prior leveraged in our SSPD framework has a finer granularity compared to previous unsupervised work.} 
    \label{different similarity}
\end{figure}

\begin{figure}[H]
    \centering
    \includegraphics[width=0.4\textwidth]{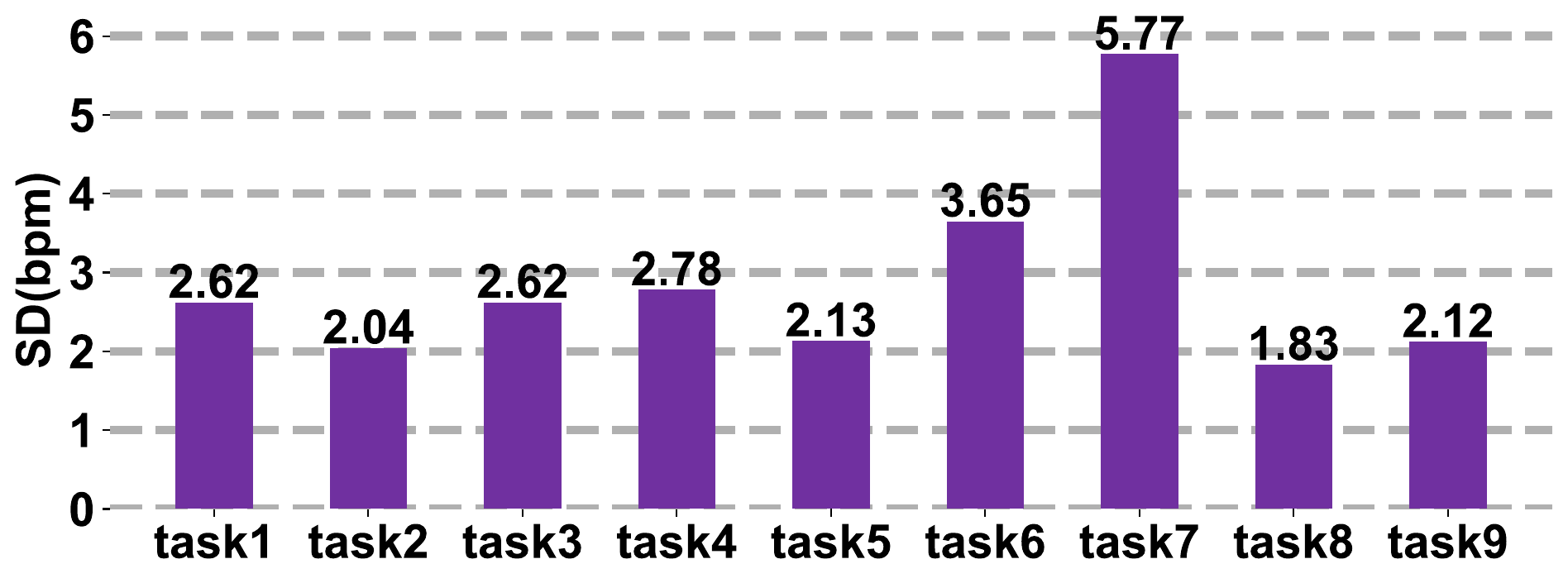}
    \caption{The mean heart rate standard deviations across different tasks in the VIPL-HR dataset, with task 7: exercise exhibiting the most significant heart rate change.} 
    \label{VIPL STD}
\end{figure}

Previous unsupervised physiological measurement methods tend to employ contrastive learning, 
aiming to minimize the distance between positive pairs in latent embeddings \cite{Self-rPPG, SLF-RPM, ViViT}, 
PSD \cite{Contrast-Phys, RemotePPG, ContrastPhys2}, or rPPG outputs \cite{yue2022}. 
Both RemotePPG \cite{RemotePPG} and Contrast-Phys \cite{Contrast-Phys} consider the temporal similarity of rPPG signals and create positive sample pairs by sampling different time windows. 
Compared to this instance-level similarity, the self-similarity within the SSPD framework introduces a much finer granularity. 
SSPD encourages the model to perceive the temporal similarity of each token to other tokens, 
aiming to capture temporal rhythms within individual time windows. For instance, in the non-stationary heart rate scenario, 
directly sampling time windows with a length of $T$/2 \cite{Contrast-Phys, RemotePPG} may result in two identical PSDs, 
as shown in Fig. \ref{different similarity}(a). Consequently, the rPPG signal predicted through these two time windows will be indistinguishable. 
However, as highlighted in gray in Fig. \ref{different similarity}(b), our SSM captures non-stationary heart rate effectively, 
aiding the model in learning more fine-grained dynamics in rPPG. 
To further demonstrate the efficacy of our self-similarity prior, we calculate the instantaneous heart rate changes of each task in the VIPL-HR dataset, 
as illustrated in Fig. \ref{VIPL STD}. In line with the experimental results in Fig. \ref{VIPL plot}, 
SSPD demonstrates the most significant performance improvement compared to Contrast-Phys in tasks with substantial heart rate changes, 
particularly in task 7: exercise and task6: long
distance \cite{VIPL}. 

\begin{figure}
    \centering
    \captionsetup[subfloat]{labelsep=none, format=plain, labelformat=empty}
    \subfloat{
        \includegraphics[width=0.15\textwidth]{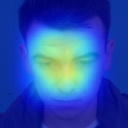}
        \includegraphics[width=0.15\textwidth]{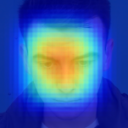}
        \includegraphics[width=0.15\textwidth]{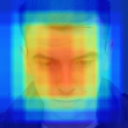}
    }\vspace{-0.18cm}

    \subfloat[(a) Saliency maps generated from different convolutional layers in the SSPD backbone on the UBFC-rPPG (top) and PURE (bottom) datasets. Each map, from left to right, corresponds to ConvBlock1\_Conv2, ConvBlock2\_Conv2, and ConvBlock3\_Conv2, respectively.]{
        \includegraphics[width=0.15\textwidth]{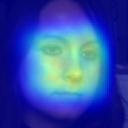}
        \includegraphics[width=0.15\textwidth]{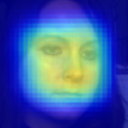}
        \includegraphics[width=0.15\textwidth]{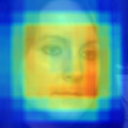}
    }

    \subfloat[(b) Saliency maps generated from three subjects in the VIPL-HR dataset using ConvBlock2\_Conv2.]{
        \includegraphics[width=0.15\textwidth]{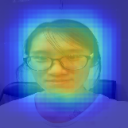}
        \includegraphics[width=0.15\textwidth]{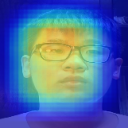}
        \includegraphics[width=0.15\textwidth]{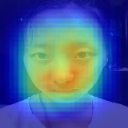}
    }
    \caption{Saliency maps generated from different convolutional layers and subjects in multiple datasets during the inference stage by SSPD.} 
    \label{CAM visualization}
\end{figure}

\subsection{Interpretability}
In Fig. \ref{CAM visualization}, we generate saliency maps during the inference stage based on LayerCAM \cite{LayerCAM}. 
Specifically, we first train our SSPD framework in an unsupervised manner. 
Subsequently, we employ BVP ground truth and the negative Pearson loss ($\mathcal{L}_{Pearson}(\cdot)$) to compute gradients for multiple convolutional layers within the backbone. 
Finally, we follow the LayerCAM to derive the gradient-weighted activation map ${M}_i \in \mathbb{R}^{T\times H_i \times W_i}$ for the i-th layer. 
Noteworthy, for visualization purposes, we average the ${M}_i$ along the temporal dimension to obtain $\widetilde{{M}_i} \in \mathbb{R}^{H_i \times W_i}$. 
Then, we combine it with the first input frame to establish spatial correspondence. 
The first two rows in Fig. \ref{CAM visualization} are generated from different convolutional layers in the UBFC-rPPG and PURE datasets. 
From left to right, they correspond to ConvBlock1\_Conv2, ConvBlock2\_Conv2, and ConvBlock3\_Conv2, respectively. 
It can be observed that as we move from shallow to deep layers, SSPD concentrates more on the facial area rather than the background, especially in the cheek and forehead regions. 
This behavior aligns with the physiological prior \cite{PhysFormer, Deepphys}. 
Moreover, the saliency maps in the last row of Fig. \ref{CAM visualization}, obtained from three subjects in the VIPL-HR dataset using ConvBlock2\_Conv2 (chosen for its favorable balance between spatial resolution and semantic information), 
exhibit the same phenomenon. SSPD consistently directs attention to areas with a high prior intensity of facial blood flow, 
enhancing its interpretability in remote physiological measurement.

\section{Conclusions and Future Work}
In this paper, we propose the Self-Similarity Prior Distillation (SSPD) framework for unsupervised remote physiological measurement, which capitalizes on the intrinsic self-similarity of cardiac activities. We first introduce a physical-prior embedded augmentation technique to mitigate the effect of various types of noise. Then, we tailor a self-similarity-aware network with a separable self-similarity model to disentangle reliable physiological features. Finally, we develop a hierarchical self-distillation paradigm for self-similarity-aware learning and rPPG signal decoupling. The experimental results demonstrate that SSPD significantly outperforms state-of-the-art unsupervised methods and serves as a strong and efficient baseline for remote physiological measurement. Our future work includes: (1) Exploring the effectiveness of spatial self-similarity for
modeling facial blood flow and achieving more precise physiological measurements. (2) Leveraging the self-distillation framework to fine-tune multiple downstream tasks, such as atrial fibrillation detection, fatigue judgment, and emotion recognition.

\bibliographystyle{IEEEtran}
\bibliography{IEEE}

\end{document}